\documentclass[10pt,journal,compsoc]{IEEEtran}
\ifCLASSOPTIONcompsoc
  \usepackage[nocompress]{cite}
\else
  \usepackage{cite}
\fi
\usepackage{amssymb}
\usepackage{xcolor} 
\usepackage{amsmath}
\usepackage{microtype}
\usepackage{graphicx}
\usepackage{multirow}
\usepackage{subfigure}
\usepackage{stfloats}
\usepackage{booktabs}
\usepackage[breaklinks=true,bookmarks=false,colorlinks]{hyperref}
\usepackage{algorithm}  
\usepackage{algorithmicx,algpseudocode}
\usepackage{stfloats}
\usepackage{xspace}
\usepackage{wrapfig}
\usepackage{lipsum}
\usepackage{setspace}
\newcommand{\eqn}[1]{Eq.~(\ref{#1})}

\newcommand{\myparagraph}[1]{\vspace{5pt}\noindent \textbf{#1}}

\newcommand{\revise}[1]{{\textcolor{black}{#1}}}

\begin{document}

\title{VideoDG: Generalizing Temporal Relations\\ in Videos to Novel Domains}

%
%
%

\author{Zhiyu~Yao,
        Yunbo~Wang,
        Jianmin~Wang,
        Philip~S.~Yu,~\IEEEmembership{Fellow,~IEEE},
        Mingsheng~Long
\IEEEcompsocitemizethanks{
\IEEEcompsocthanksitem The authors are with the School of Software, BNRist, Tsinghua University, Beijing 100084, China.
\IEEEcompsocthanksitem Yunbo~Wang is now with the MoE Key Lab of AI, AI Institute, Shanghai Jiao Tong University, Shanghai 200240, China.
\IEEEcompsocthanksitem Zhiyu~Yao and Yunbo~Wang contributed equally to this work.
\IEEEcompsocthanksitem Corresponding author: Mingsheng Long, mingsheng@tsinghua.edu.cn.
}
}

\markboth{IEEE Transactions on Pattern Analysis and Machine Intelligence,~Vol.~XX, No.~X}%
{Yao \MakeLowercase{\textit{et al.}}: VideoDG: Generalizing Temporal Relations in Videos to Novel Domains}

\IEEEtitleabstractindextext{%

\begin{abstract}
This paper introduces video domain generalization where most video classification networks degenerate due to the lack of exposure to the target domains of divergent distributions. We observe that the global temporal features are less generalizable, due to the temporal domain shift that videos from other unseen domains may have an unexpected absence or misalignment of the temporal relations. This finding has motivated us to solve video domain generalization by effectively learning the local-relation features of different timescales that are more generalizable, and exploiting them along with the global-relation features to maintain the discriminability. This paper presents the VideoDG framework with two technical contributions. The first is a new deep architecture named the Adversarial Pyramid Network, which improves the generalizability of video features by capturing the local-relation, global-relation, and cross-relation features progressively. On the basis of pyramid features, the second contribution is a new and robust approach of adversarial data augmentation that can bridge different video domains by improving the diversity and quality of augmented data. We construct three video domain generalization benchmarks in which domains are divided according to different datasets, different consequences of actions, or different camera views, respectively. VideoDG consistently outperforms the combinations of previous video classification models and existing domain generalization methods on all benchmarks. 

\end{abstract}

}

\maketitle

\IEEEdisplaynontitleabstractindextext

\IEEEpeerreviewmaketitle

\IEEEraisesectionheading{\section{Introduction}\label{sec:introduction}}

Previous deep networks for video classification achieve competitive results in the intra-domain setting \cite{Wang16,wang2017spatiotemporal,carreira2017quo,NonLocal2018,zhou2018temporal}, assuming that training and test videos are independently and identically distributed (i.i.d.). We find that the performance of these models degenerates since the i.i.d. assumption might be violated as the domain shift is on. In this paper, we name this inter-domain application setting as the \emph{video domain generalization} problem, where models are trained on a source domain and evaluated on \emph{unseen} target domains with the same label set.  
Unlike the extensively studied image and video domain adaptation problem \cite{yosinski2014transferable,Ghifary2015domain,tzeng2017adversarial,long2018conditional,chen2019temporal,Jamal2018DeepDA}, in our setting, not only labels but also data of the target domains are unavailable during training. 
Further, learning more generalizable models is crucial for action recognition. This is because the actions from the test domain are typically performed by new subjects in unknown environments or novel views. 
As described by Zhou \textit{et al.} \cite{zhou2021domain}, intuitively, different persons can perform the same action in dramatically different ways, so, commonly, a model might not be able to recognize actions performed by new subjects not seen during training. 

\begin{figure*}
  \centering
  \subfigure[Feature discrepancy]{
    \label{fig:intro_distance} 
     \includegraphics[height=0.7\columnwidth]{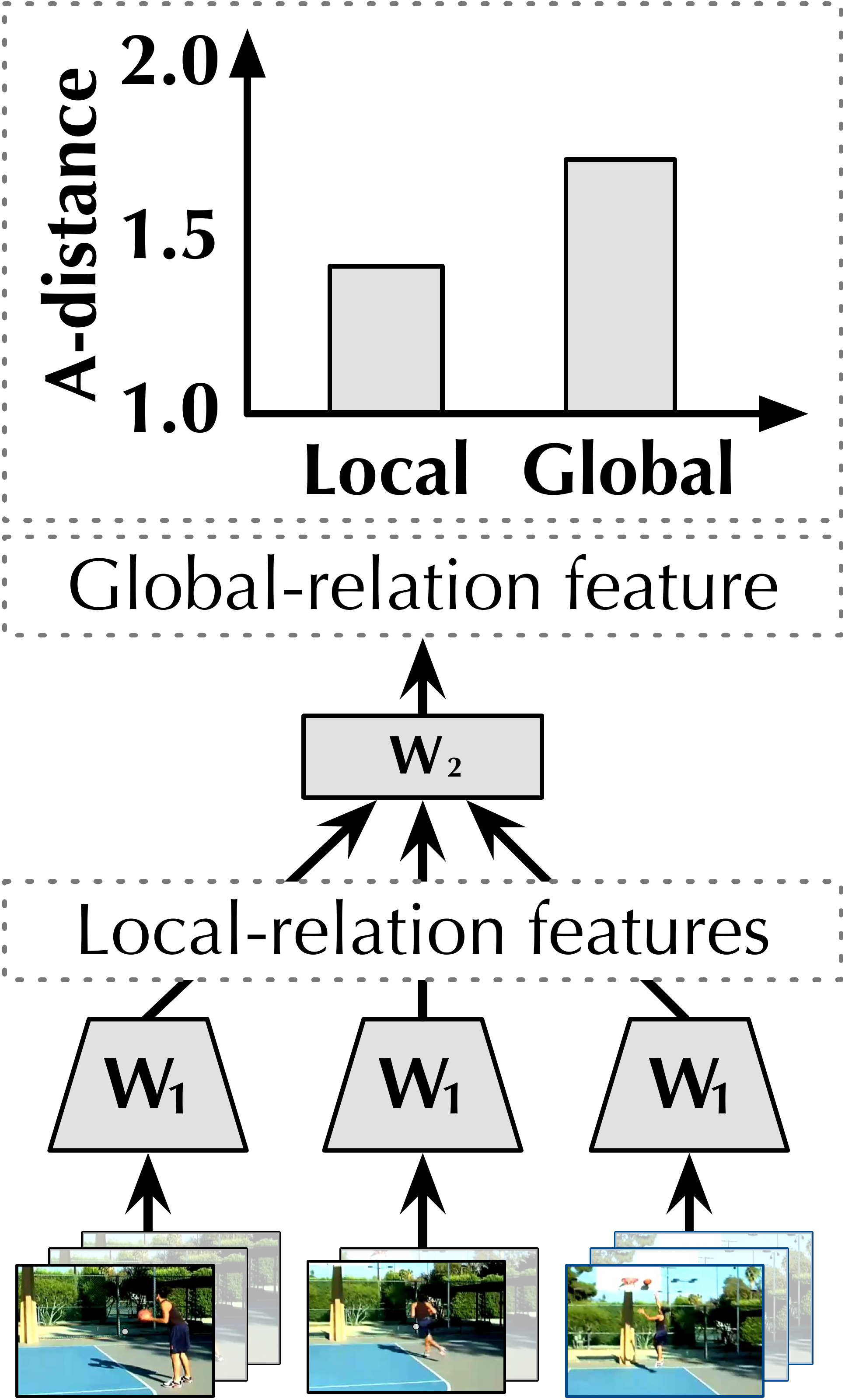}}
  \subfigure[A negative and a positive example of video domain generalization]{
  \label{fig:intro_show}
    \includegraphics[height=0.7\columnwidth]{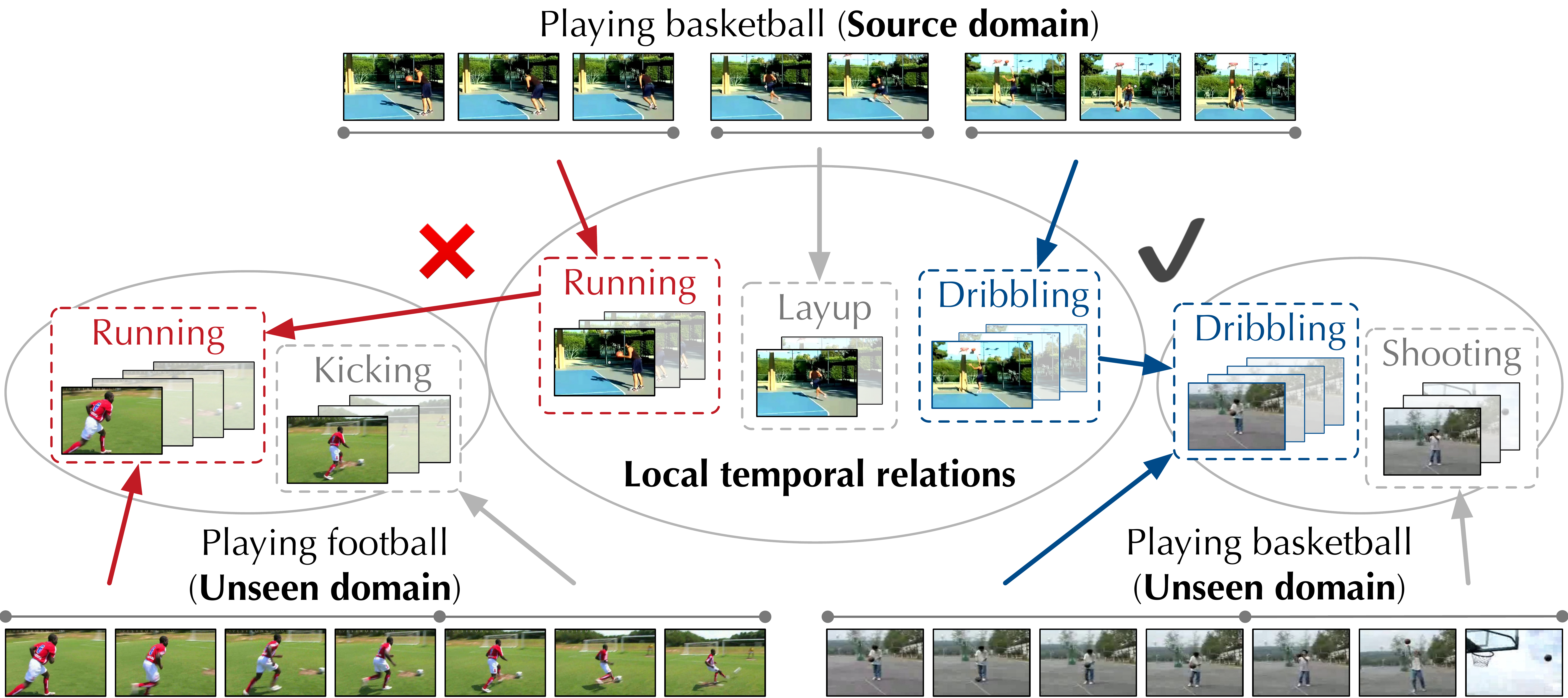}}
    \vspace{-5pt}
 \caption{Our observations of video domain generalization: (a) Local temporal relations are more generalizable than the global ones by using A-distances as a measurement of domain shift; (b) Solving video domain generalization depends on aligning the distributions of local temporal relations correctly, which can be guided by the global relations that are more discriminative in the long term. In this case, the model should mitigate the domain shift via the sub-action of \textit{dribbling} rather than \textit{running}. It is only possible by leveraging the global relation of \textit{playing basketball} as the generalization bridge.
 }
  \label{fig:subfig} 
  \label{video-dg-examples}
\end{figure*}

For the video domain generalization problem, one might ask what causes the drop in performance of the mainstream deep networks. The answer is that when the distribution of target data is unknown, \emph{the spatial and temporal domain shifts coexist}. 
The spatial domain shift is caused by the variations of the appearance in video frames, which can be partly solved by previous image domain generalization methods \cite{ADA,Jigsaw,dou2019domain}.
The temporal domain shift indicates the
unexpected absence or misalignment of short-term video events (\emph{a.k.a.} local temporal relations) across distant domains, which has not been fully considered by existing approaches.
As shown in Fig.~\ref{fig:intro_show}, videos in different domains under the same category (\textit{i.e.}, \textit{playing basketball}) may have distinct subsets or sequences of local events (\textit{e.g.}, running $\rightarrow$ layup $\rightarrow$ dribbling vs. dribbling $\rightarrow$ shooting), which presents a new challenge to learning diverse features for classification that can cover various permutations and combinations of basic local video relations.

\subsection{To Solve the Dilemma of Generalizability against Discriminability}

Recent approaches like Temporal Relation Networks (TRN) \cite{zhou2018temporal} learn hierarchical features over long-term video clips, with lower layers focusing on local temporal relations and higher layers focusing on more global ones. 
Throughout this paper, we borrow the A-distance  \cite{Ben-DavidBCKPV10} to quantify the overall domain shift, which measures the distance between two separate domains with respect to the same feature mapping function.
In an early experiment shown in Fig.~\ref{fig:intro_distance}, we used the TRN learned in the UCF domain \cite{UCF101} as the feature mapping function, and tested its performance for the HMDB domain \cite{HMDB51}. We observed that the local-relation features yield narrow A-distance \cite{Ben-DavidBCKPV10} (the lower, the better), suggesting that while the local-relation features learned from short video snippets are less discriminative in intra-domain scenarios, they are more generalizable than the global-relation features.
Based on this finding, our solution to the temporal domain shift is to use the short-term video relations to generate adversarial examples to augment the source dataset.
The motivation is that although the local video events under the same category might not be identical in different domains, there must be some overlap between them that can be used as generalization bridges.
As shown in Fig.~\ref{fig:intro_show} (bottom right), the model can recognize a video of \textit{playing basketball} from the unseen domain via a shared local event of \textit{dribbling}.

However, there comes another problem. 
We observe that generally the above domain generalization approach deteriorates in-domain performance.
For example, the model may misrecognize a video of \textit{playing football} (bottom left) as \textit{playing basketball} via the shared local event of \textit{running}. The fact is that while the use of local-relation features to generate new adversarial examples can improve the generalizability of deep networks, it may also degrade the discriminability of spatiotemporal features, resulting in videos being misclassified into similar categories.
The key idea is to use the global-relation features to guide the generalization of local events and dynamically find the events that are highly relevant to the overall video representation.
Specifically, we present the Adversarial Pyramid Network (APN), which corrects the erroneous generalization with two techniques. First, it aligns the local- and global-relation features by performing a pyramid of attention blocks at different timescales. In the process of domain augmentation, the use of the multi-level relational features can balance the diversity and the focus of the generated data points.
Second, we introduce a new training algorithm of Robust Adversarial Domain Augmentation (RADA), which improves the robustness of APN to various adversarial perturbations derived from the relational features at multiple levels.
Our methods are inspired by several cutting-edge techniques from other ML topics, but more importantly, we improve them, \textit{i.e.}, ADA \cite{ADA} $\rightarrow$ RADA, Transformer \cite{Transformer} $\rightarrow$ APN, and jointly use them with strong ties in a unified framework named \textbf{VideoDG}.

We design three benchmarks for video domain generalization, covering different transfer learning scenarios across datasets, across camera views, and across different consequences of video events. The experiments also cover different domain generalization setups with multiple source domains or multiple target domains. These setups are well-explored for image data, but have never been studied in the spatiotemporal video context.
VideoDG consistently achieves the best results on all benchmarks.
We release the code and pre-trained models at \url{https://github.com/thuml/VideoDG}.

To sum up, the contributions of this paper are as follows:
\begin{itemize}
    \item It introduces video domain generalization, a new transfer learning problem that has not been explored before. It is also very challenging due to the co-occurrence of spatial and temporal domain shifts, and cannot be well resolved by existing video classification models, nor the na\"{i}ve modifications of image domain generalization approaches to video data.
    \item We find that the key to video domain generalization is to resolve the dilemma of generalizability and discriminability. The main idea is to expand the frame relations of the source domain properly, in the sense that they need to be diverse enough to be easily generalized to potential target domains, while being discriminative to be correctly recognized.
    \item 
    We also observe that, empirically, in existing video classification models, local relational features are more generalizable while the global ones are more discriminative. It enlightens us to propose the APN model and the RADA algorithm to form a unified framework, VideoDG, in which features at different pyramid levels play different roles in the process of robust adversarial domain augmentation. 
\end{itemize}

\section{Related Work}

\subsection{Video Classification}
Recent advances in deep learning provide some useful insights on how to capture long-term temporal relations to improve the discriminability of spatiotemporal features \cite{martinez2019action, DynamoNet, SlowFast, Long-term}. 
However, most existing video classification models, including 2D CNNs \cite{Simonyan14,Karpathy14,feichtenhofer2016spatiotemporal,Wang16,zhou2018temporal, sun2017optical, STM, lin2019tsm}, 3D CNNs \cite{Tran15,carreira2017quo,qiu2017learning,xie2017rethinking, CSN, yang2020temporal}, and attention-based models \cite{attention-pooling,NonLocal2018,zhang2018attention,wang2018eidetic}, do not consider non-i.i.d. learning settings, which may lack sufficient generalizability for unseen domains.  In contrast, our approach takes into account the unexpected absence or permutation of video events, \textit{i.e.}, temporal domain shift, and seeks a balance between the ability to distinguish and the ability to generalize.
A significant difference between the existing video domain adaptation models \cite{Jamal2018DeepDA,chen2019temporal} and our approach is that the target domain is not accessible in this paper. In other words, the video domain generalization problem is more challenging and not yet well explored.

Among all these models mentioned above, our APN model is most related to the Temporal Relation Network (TRN) \cite{zhou2018temporal}, which learns local temporal features from different lengths and different combinations of short video snippets and then ensembles them together to get a global sequence-level feature. However, we find that the performance of TRN deteriorates in the domain generalization setting. We also find that TRN cannot further benefit from the modern image domain generalization method \cite{ADA}. 

Our work is distinct from TRN in two perspectives. First, our approach is fully motivated from the view of domain generalization. Thus we consider the cross-relation features in addition to the local relational features to trade off the discriminability and transferability. Second, we introduce a new approach that effectively adapts the previous ADA method to video domain generalization. Different from TRN, our approach leverages a pyramid architecture based on the Transformer self-attention mechanism \cite{Transformer} to progressively fuse temporal relations at different time scales. From this perspective, our work is also related to recognition models that incorporate attention blocks \cite{attention-pooling,NonLocal2018,zhang2018attention,wang2018eidetic}. Our model organizes the attention blocks in a pyramid framework, which is driven by a clear motivation and validated by extensive experiments

\subsection{Image Domain Generalization} 
Most previous domain generalization methods are designed for image data \cite{li2018domain,li2019feature,shankar2018generalizing,ADA,Jigsaw,dou2019domain, li2018DG,qiao2020learning,shu2021open}, which can be divided into two groups: feature-based methods and data-based methods. Feature-based methods focus on extracting invariant cross-domain representations.
Li \textit{et al.} \cite{li2018domain} introduced an adversarial learning approach based on the Maximum Mean Discrepancy (MMD). A meta-learning approach \cite{li2019feature} was proposed to train a domain-invariant feature extractor. 
Li \textit{et al.} \cite{li2019episodic} improved the robustness of the deep network to unseen image domains with a new episodic training strategy that can mimic the train-test domain shift.

Data-based methods connect the source domain distribution and the unseen target distribution by expanding the training dataset. 
Volpi and Murino \cite{volpi2019addressing} and Jackson \textit{et al.} \cite{jackson2019style} showed that some specific data augmentation methods, such as brightness perturbations, color jittering, and rotations, can improve the generalization performance for image classification.
However, there is no strong evidence for the effectiveness of these simple operations on video domain generalization; and we will do some experimental comparisons.
Other techniques augment the source domain with gradient-based perturbations \cite{shankar2018generalizing,sinha2017certifying,ADA}.
Our approach is closely related to the work from Sinha \textit{et al.} \cite{sinha2017certifying} and Volpi \textit{et al.} \cite{ADA}; the first one introduced a new surrogate loss based on Wasserstein distance for generating adversarial examples on the fly at training time, while the second one, namely the Adaptive Data Augmentation (ADA) method, appends the training set with the adversarial examples over iterations.
Unlike all the above models, our approach is an early work for video domain generalization, which extends the basic ADA method by particularly considering how to mitigate the co-occurring spatial and temporal domain shifts.

\subsection{Transfer Learning for Video Data}

Domain adaptation is a classic problem in the field of transfer learning \cite{yosinski2014transferable,Ghifary2015domain,tzeng2017adversarial,long2018conditional}, in which, different from the domain generalization setups, label-free data samples from the target domain are available during training.
Some recent approaches were proposed to solve the domain adaptation problem for videos \cite{Jamal2018DeepDA,chen2019temporal, pan2020adversarial, munro2020multi}. They are related to our work because the temporal domain shift exists in both video domain adaptation and generalization problems.
However, as mentioned above, these models learn domain-invariant features by harnessing the data samples from the target domain, and thus cannot be directly used for video domain generalization. In this work, our goal is to learn more generalizable representations purely from the source videos, without accessing any clues of the target domain.

\section{Preliminaries}

The proposed method, VideoDG, has two technical contributions. The first one is a new network architecture named Adversarial Pyramid Network (APN), which considers both the generalizability and discriminability of spatiotemporal features.
The second one is a new adversarial domain augmentation algorithm that is specifically designed for video domain generalization. It improves the previous Adaptive Data Augmentation (ADA) method from both pyramid and robustness aspects. 

\subsection{Adaptive Data Augmentation}

Sinha \textit{et al.} \cite{sinha2017certifying} proposed an adversarial training procedure, named WRM, to obtain a robust neural network for image perturbations, which augments the model
parameter updates by perturbing the underlying
distribution of training data in a Wasserstein ball.
Based on WRM, the Adaptive Data Augmentation (ADA) method \cite{ADA} casts image domain generalization as the following worst-case problem around the source distribution $Q$ and is expected to generalize well on unseen domains $P$: 
\begin{equation}
\label{equal:worst-case}
    \underset{\theta \in \Theta}{\operatorname{min}} \sup _{P : D(P, Q) \leq d} \mathbf{E}_{P}\big[\ell_\theta \left(X, Y\right)\big],
\end{equation}
where $\theta \in \Theta$ is the set of weights of the entire model. $(X, Y) \in \mathcal{X} \times \mathcal{Y}$ indicates a source data point with its label. $\ell: \mathcal{X} \times \mathcal{Y} \rightarrow \mathbf{R}$ is the categorical cross-entropy loss.
The ADA method denotes by $D(P, Q)$ the distance metric around the source distribution $Q$ that characterizes the set of unknown populations we wish to generalize to. The perturbed new data distribution $P$ should be diverse enough but not deviate far from $Q$, \textit{i.e.}, $D\left(P, Q\right) \leq d$. $D(P, Q)$ is defined by the Wasserstein distance on the semantic space. Consider the transportation cost from two data points $(X, Y)$ to $\left(X^{\prime}, Y^{\prime}\right)$: 
\begin{equation}
c\left((X, Y),\left(X^{\prime}, Y^{\prime}\right)\right) \triangleq \frac{1}{2}\left\|X-X^{\prime}\right\|_{2}^{2}+\infty \cdot {\mathbf{1}}\left\{Y \neq Y^{\prime}\right\}, 
\end{equation}
which is denoted by $c\left(X,X^{\prime}\right)$ if $Y=Y^\prime$. 
By taking $g(X)$ as the output of the last hidden layer, the distance of two data points in the original space $\mathcal{X} \times \mathcal{Y}$ is defined as:
\begin{equation}
\label{max-phase}
    c_{\theta}\left((X, Y),\left(X^{\prime}, Y^{\prime}\right)\right)=c\left(\left(g(X), Y\right), \left(g(X^{\prime}), Y^{\prime}\right)\right),
\end{equation}
in which $c_{\theta}$ measures distance with respect to output of the last hidden layer.
ADA uses the semantic space constraint.
The worst-case formulation can be defined as a surrogate loss:
\begin{equation}
    \label{max-mization-problem}
    \sup _{X \in \mathcal{X}}\big\{\ell_\theta \left(X, Y_{0}\right)-\gamma c_{\theta}\left((X, Y_{0}),(X_{0}, Y_{0})\right)\big\},
\end{equation}
where $\gamma$ is a hyperparameter of the transportation cost and $(X_0, Y_0)$ is a data point from the source distribution $Q$. 
The training procedure of ADA has two separate stages: a data augmentation stage and a minimization stage with respect to $\ell_\theta$. The data augmentation stage has two alternated training phases: a maximization phase with respect to \eqn{max-mization-problem} and an online minimization phase of $\ell_\theta$ on the augmented dataset. In the maximization phase, the new data point $(X_k, Y_0)$ is generated iteratively to learn ``hard'' data points from fictitious target distributions by maximizing the following perturbation over the source data $X_0$:
\begin{equation}
\label{max-mization}
    {X} \leftarrow {X} + \nabla_{X} \big \{\ell_\theta \left(X, Y_0\right) - \gamma c_{\theta} \left((X, Y_0), (X_0, Y_0)\right) \big \}.
\end{equation}

However, the original image ADA method is not suitable for video domain generalization in two ways.
First, it defines the transportation cost $c(\cdot)$ at the activation of the last hidden layer, which could be less generalizable on video data. By contrast, we define $c(\cdot)$ at different network levels to trade-off the generalizability of the local temporal features and the discriminability of the global temporal features, leading to more diverse adversarial examples.
Second, the robustness of the original ADA method is not well guaranteed because the augmented data may be too diverse.

\subsection{Combining ADA and Video Classification Models}

Alg. \ref{alg:baseline-Framwork} shows a simple baseline approach that slightly modifies the original ADA to fit the existing video classification models \cite{Wang16,lin2019tsm,zhou2018temporal,carreira2017quo,NonLocal2018}. 
Notably, the original ADA method has two separate training stages (we refer to the proposed algorithm in \cite{ADA}): it first generates new images in $K$ minimax training procedures to augment the source dataset $K$ times, and subsequently learns the model by minimizing the classification loss over the entire augmented dataset. 
However, such a two-stage training strategy is not suitable for video data. 
One concern is the memory footprint, because as $K$ grows, the amount of video data that needs to be stored increases dramatically.
Moreover, to learn the semantic relations in different timescales and avoid over-fitting, video classification models commonly take as inputs different combinations of frames that are sampled randomly or uniformly from the entire videos.
But with the growth of $K$, in the second training stage, the original ADA method is more and more likely to sample the adversarial examples with fixed anchor frames, which would harm the diversity of input data.
To solve the above problems, we combine the two separate training stages of image ADA, generating adversarial examples and optimizing the classification loss over the augmented data \textbf{on-the-fly} without appending the training set ({Line 11 in Alg. \ref{alg:baseline-Framwork}}).
Following \eqn{max-mization}, we use the activations of the last hidden layer to compute the transportation cost $c(\cdot)$ in  $T_\text{max}$ maximization phases. 
However, in spatiotemporal scenarios, in order to generate both diverse and representative adversarial examples, we need to specifically consider the temporal relations at different timescales. It encourages us to trade-off the generalizability of the local temporal features and the discriminability of the global temporal features. In Section \ref{sec:video_ada}, we give detailed descriptions of RADA, which can be viewed as an improved version of ADA in the video context.

\begin{algorithm}[t] 
\setstretch{1.1}
  \caption{Applying a memory-efficient ADA baseline to TSN \cite{Wang16}, TSM \cite{lin2019tsm}, TRN \cite{zhou2018temporal}, I3D \cite{carreira2017quo}, and NL-I3D \cite{NonLocal2018}}  
  \label{alg:baseline-Framwork}  
  \begin{algorithmic}[1]
    \Require
    A source video dataset $\mathcal{S} = \left\{({X}_{i}, {Y}_{i})\right\}_{i=1}^n$, the number of maximization phases $T_\text{max}$, the penalty parameter of the transportation cost $\gamma$, the learning rate $\alpha$
    \Ensure  
    Learned network weights $\theta$
    \State \textbf{Initialize} $\theta \leftarrow \theta_{0}$
    \Repeat
        \State // Randomly sample a batch of data
        \State $({X}, {Y}) \sim \mathcal{S}$
        \State // Minimize the cls. loss of the source data
        \State $\theta \leftarrow \theta-\alpha \nabla_{\theta} \ell_{\theta}\left(X, {Y}\right)$
        %
        \State ${X}^\textrm{adv} \leftarrow {X}$
        \For{$t=1, \ldots, {T}_\textrm{max}$}
            \State // Generate adversarial examples
            \State $\ell^{\text{surrogate}} = \ell_{\theta}\left(X, {Y}\right) -  \gamma c_{\theta} \left(\left(X^\textrm{adv}, {Y}\right), \left(X, Y \right) \right)$
            \State ${X}^\textrm{adv} \leftarrow {X}^\textrm{adv} + \nabla_{X} \ell^{\text{surrogate}}$
        \EndFor
        \State // Minimize the cls. loss of adversarial examples
        \State $ \theta \leftarrow \theta-\alpha \nabla_{\theta} \ell_{\theta}\left(X^\textrm{adv}, {Y}\right)$
    \Until{Convergence}
  \end{algorithmic}  
\end{algorithm}

%

\section{Method}

Our method, VideoDG, has two major technical contributions. In this section, we first introduce the basic building blocks and overall architecture of the Adversarial Pyramid Network (APN) as shown in Fig.~\ref{fig:apn}. Then, we propose a new training approach of Robust Adversarial Domain Augmentation (RADA) to improve the generalizability of APN.

\subsection{Basic Building Blocks}

\subsubsection{Frame Encoder} 

Given an input video, we divide it evenly into $M$ segments and then randomly sample one frame from each segment. We extract a $D$-dimensional feature $f_i$ ($D=256$) from each selected frame using a ResNet-$50$ \cite{HeRes}, a fully-connected layer, and a Dropout layer. 
We then concatenate $\{{f_1}, \ldots ,{f_M}\}$ at the time dimension and obtain $\mathcal{F}_{M} \in {R}^{M \times D}$ shown in Fig.~\ref{fig:apn}.

\subsubsection{Attention Block} 

\begin{figure*}[t]
    \centering
    \includegraphics[width=\textwidth]{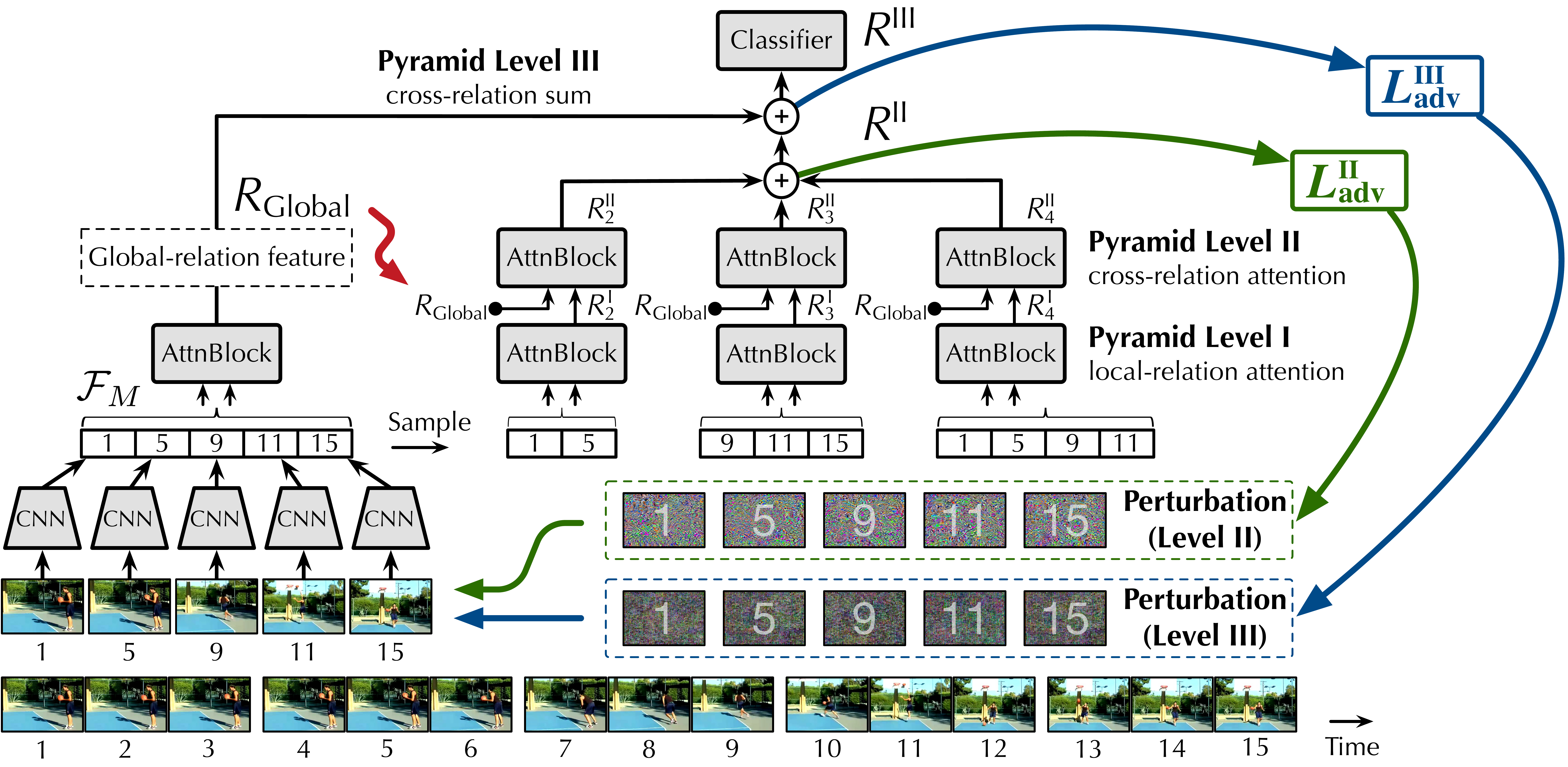}
    \vspace{-20pt}
    \caption{
    An overview of \textbf{VideoDG}, which first uses the APN model to extract local-relation, global-relation, and multilayer cross-relation features progressively. It then uses the cross-relation features to generate spatiotemporal adversarial examples, compromising between generalizability and discriminability. VideoDG tries to mitigate the temporal domain shift from both the perspectives of representation learning and data augmentation.}
    \label{fig:apn}
\end{figure*}

APN contains multiple Transformer-style blocks \cite{Transformer} organized in a pyramid. Each block has $3$ multi-head attention layers, $2$ fully-connected layers, and layer normalization \cite{LN}. The multi-head attention layer takes queries, keys, and values as inputs: $\text{Multihead-Attention}(\textrm{Query}, \textrm{Key} , \textrm{Key})$, which we share common inputs for keys and values. This block is simply represented by: $\texttt{AttnBlock}(\textrm{Query}, \textrm{Key})$, while we remove the position embedding module and add a Dropout layer in the end, as shown in Fig.~\ref{fig:apn}.

\subsection{Adversarial Pyramid Network}

The idea of using multi-scale temporal relations is initially inspired by TRN \cite{zhou2018temporal}.
However, unlike TRN, we innovatively propose a pyramid network of relational features based on the above-mentioned attention blocks to gradually learn the local- and cross-relation features on different timescales, which is essential for alleviating the temporal domain shift (\textit{i.e.}, absence or misalignment of local video events). Another benefit of the pyramid network is that it can greatly facilitate the process of domain augmentation by balancing the diversity and representativeness of new data points.

\subsubsection{Pyramid Level I: Local-Relation Attention} 
The first pyramid level is applied to the output of the frame encoder.  
For each $m \in\{2, \ldots, M-1\}$, it first randomly samples $m$ {consecutive} features from $\{{f_1}, \ldots, {f_M}\}$, and then concatenates them at the time dimension to obtain $\mathcal{F}_{m}$.
Next, it uses the attention block to get features of Pyramid Level I:
\begin{equation}
    \mathcal{R}_{m}^\textrm{I} =
\texttt{AttnBlock}(\mathcal{F}_{m}, \mathcal{F}_{m}),
\end{equation}
and provides a set of local-relation features $\{{ \mathcal{R}_{2}^\textrm{I}}, \ldots ,{ \mathcal{R}_{M-1}^\textrm{I}}\}$, which can represent a variety of short-term video events.

\subsubsection{Pyramid Level II: Cross-Relation Attention} 
When we review the showcases of \textit{playing basketball} vs. \textit{playing football} in Fig.~\ref{video-dg-examples}, we find that local temporal cues can be more generalizable but may also lead to false generalization. 
To avoid generating unnecessary adversarial examples that diverge too much, we cannot directly use $\mathcal{R}_{m}^\textrm{I}$ for domain augmentation. Instead, we need to constrain the spatiotemporal features using more category-specific information.
To this end, we use the second pyramid level to align each local-relation feature $\mathcal{R}_{m}^\textrm{I} \in \{{\mathcal{R}_{2}^\textrm{I}}, \ldots ,{\mathcal{R}_{M-1}^\textrm{I}}\}$ to the global one
\begin{equation}
    \mathcal{R}_{\text{Global}} =\texttt{AttnBlock}(\mathcal{F}_M, \mathcal{F}_M)
\end{equation}
which can be more discriminative by considering long-term video cues.
Concretely, we perform the attention block to enable the temporal relations at various timescales to interact with each other:
\begin{equation}
    \mathcal{R}_{m}^\textrm{II} =\texttt{AttnBlock}(\mathcal{R}_\text{Global}, \mathcal{R}_{m}^\textrm{I}),
\end{equation}
by taking $\mathcal{R}_\text{Global} \in {R}^{M \times D}$ as the query and $\mathcal{R}_{m}^\textrm{I} \in {R}^{m \times D}$ as the key.
Finally, to combine cross-relation features at different timescales at Pyramid Level II, we have $\mathcal{R}^\textrm{II} = \sum_{m=2}^{M-1}{\mathcal{R}_{m}^\textrm{II}}$ and use it as a part of the RADA algorithm to be discussed later. Note that we also explored other aggregation functions, \textit{e.g.}, concatenation and attention, but the element-wise sum empirically performs best.

\subsubsection{Pyramid Level III: Cross-Relation Sum} 

Although the cross-relation attention block at the second pyramid level enables the local-relation features to interact with the global one, it maintains the strong impact of $\mathcal{R}_{m}^\textrm{I}$ by taking it as both key and value, rather than directly reflecting the content of the global-relation feature. 
Therefore, at the third pyramid level, we aggregate  $\mathcal{R}^\textrm{II}$ and  $\mathcal{R}_\text{Global}$ by $\mathcal{R}^\textrm{III} = \mathcal{R}^\textrm{II} + \mathcal{R}_\text{Global}$. 
Empirically, we explored other aggregation functions such as concatenation and attention, and found that the element-wise sum is most effective. 
At test time, we use $\mathcal{R}^\textrm{III}$ for classification, while at training time, we use both $\mathcal{R}^\textrm{II}$ and  $\mathcal{R}^\textrm{III}$ in the process of domain augmentation, as they are complementary to each other, resulting in both diverse and representative adversarial examples. In Section \ref{sec:expri}, we show the necessity of using hierarchical features in RADA.

To sum up, APN enables learning a pyramid of relational features. Potentially, each level of features can play a different role in the process of generating adversarial examples: 
(1) $\mathcal{R}_m^\textrm{I}$ is concerned with generalizability rather than discriminability, which can enhance the diversity of learned features; 
(2) $\mathcal{R}^\textrm{II}$ adaptively focuses on category-specific local relations with the help of cross-relation attention, which can be used to expand the distribution of the generated spatiotemporal data; 
(3) $\mathcal{R}^\textrm{III}$ directly shows the discriminative long-term relations to the classifier and the RADA algorithm, which can be used to control the focus of the expanded source domain.


\subsection{RADA Algorithm}
\label{sec:video_ada}

To use APN as a solution to video domain generalization, a straightforward method is to directly combine it with existing methods for learning generalizable visual features, \textit{e.g.} Adaptive Data Augmentation (ADA) \cite{ADA}. 
The adversarial examples can be viewed as difficult data points far from the source domain. They can be close to the target domain, or at least they can enable the model to learn more generalizable features from the augmented training set.

However, the vanilla ADA method does not consider the specific challenge of video domain generalization, \textit{i.e.}, the temporal domain shift.
In the first place, we need the augmented data points to be diverse enough to cover a variety of absences and permutations of local video events across domains.
In the second place, we want the distribution of the generated data to be as close as possible to that of the invisible target domain.
To balance the diversity and focus of new data points and to learn robust, domain-invariant representations, we propose a new algorithm named Robust Adversarial Domain Augmentation (RADA) based on the APN model as the training framework of VideoDG. 
RADA inherits the theory of ADA and improves it for video scenarios by leveraging the relational features of multiple levels and the robustness regularization.
As shown in Alg. \ref{alg:Framwork}, RADA consists of two training procedures:
\begin{itemize}
    \item $T_\text{max}$ maximization phases to generate adversarial examples from the above-mentioned multi-level relational features, which control the distribution of the expanded source domain.
    \item A minimization phase of classification errors with a \textit{robustness regularization}, which allows the learning of generalizable features to be unaffected by overly divergent new data points.
    \vspace{-5pt}
\end{itemize}

\begin{algorithm*}
\setstretch{1.1}
  \caption{The RADA framework for training APN}  
  \label{alg:Framwork}  
  \begin{algorithmic}[1]
    \Require  
    A source video dataset $\mathcal{S} = \left\{({X}_{i}, {Y}_{i})\right\}_{i=1}^n$, the penalty parameter of the transportation cost $\gamma$, the robustness regularization parameter $\lambda$, the number of maximization phases $T_\text{max}$, and the learning rate $\alpha$
    
    \Ensure  
    Learned APN weights $\theta$
    \State \textbf{Initialize} $\theta \leftarrow \theta_{0}$
    \Repeat
        \State $({X}, {Y}) \sim \mathcal{S}$
        \Comment{Randomly sample a batch of data}
        \State $(\mathcal{R}^\textrm{II}_{0}, \mathcal{R}^\textrm{III}_{0}) = \mathrm{APN}(X)$
        \State $\theta \leftarrow \theta-\alpha \nabla_{\theta} \ell\left(h(\mathcal{R}^\textrm{III}_{0}), {Y}\right)$
        \Comment{Minimize the classification loss of the source data}
        \State ${X}^\textrm{II} = {X}; {X}^\textrm{III} = {X}$
        \For{$t=1, \ldots, {T}_\text{max}$}
        \Comment{For each maximization phase}
            \State $(\mathcal{R}^\textrm{II}, \underline\quad) = \mathrm{APN}(X^\textrm{II})$
            \State ${X}^\textrm{II} \leftarrow {X}^\textrm{II} +  \nabla_{X} L_{\textrm{adv}}^\textrm{II} \left(\theta ; (X,Y)\right)$
            \Comment{Generate new data according to Eq. \eqref{pyramid-II-loss}}
            \State $(\underline\quad, \mathcal{R}^\textrm{III}) = \mathrm{APN}(X^\textrm{III})$
            \State ${X}^\textrm{III} \leftarrow {X}^\textrm{III} + \nabla_{X} L_{\textrm{adv}}^\textrm{III} \left(\theta ; (X,Y)\right)$
        \EndFor
        
        \State $(\mathcal{R}^\textrm{II}, \underline\quad) = \mathrm{APN}(X^\textrm{II})$ 
        \Comment{For minimization phase with robust training}
        
        \State $(\underline\quad, \mathcal{R}^\textrm{III}) = \mathrm{APN}(X^\textrm{III})$
        \State $ \theta \leftarrow \theta-\alpha \nabla_{\theta} (L_{\textrm{cls}}^\textrm{II}(\theta ; ({X}^\textrm{II},Y)) + \lambda L_{\textrm{robust}}^\textrm{II}(\theta ; ({X}^\textrm{II},Y)))$
        \Comment{According to Eq. \eqref{robust-loss}}
        
        \State $ \theta \leftarrow \theta-\alpha \nabla_{\theta} (L_{\textrm{cls}}^\textrm{III}(\theta ; ({X}^\textrm{III},Y)) + \lambda L_{\textrm{robust}}^\textrm{III}(\theta ; ({X}^\textrm{III},Y)))$
    \Until{Convergence}
  \end{algorithmic}  
\end{algorithm*}

\subsubsection{Maximization Phases with Feature Pyramid}
We use hierarchical cross-relation features, ${\mathcal{R}}^\textrm{II}$ and ${\mathcal{R}}^\textrm{III}$, to generate data perturbations, rather than the local- or global-relation features, \textit{i.e.}, $\{\mathcal{R}_m^\textrm{I}\}_{m=2}^{M-1}$ and $\mathcal{R}_\text{Global}$.
Though both $\mathcal{R}^\textrm{II}$ and $\mathcal{R}^\textrm{III}$ can balance the generalizability and discriminability of the spatiotemporal features, $\mathcal{R}^\textrm{II}$ is a direct reflection of $\{\mathcal{R}_m^\textrm{I}\}_{m=2}^{M-1}$, while $\mathcal{R}^\textrm{III}$ contains more category-specific knowledge from $\mathcal{R}_\text{Global}$.
Hence, using both of them as the roots to generate new data points can control the distribution of the expanded source domain.
Based on different levels of features in APN, we generate adversarial examples $X^\text{II}$ and $X^\text{III}$, and update them iteratively in $T_\text{max}$ phases shown in Lines 7-12 in Alg. \ref{alg:Framwork}. In each phase, we extend the original image-based ADA method
to hierarchical, spatiotemporal feature space and maximize:
\begin{equation}
    \label{pyramid-II-loss}
    L_{\textrm{adv}}^k \left(\theta ; (X,Y)\right) := \sup_{X \in \mathcal{X}} \left\{\ell\left(h(\mathcal{R}^k), Y\right) -\gamma c \left(\mathcal{R}^k, \mathcal{R}^k_{0} \right)\right\},
\end{equation}
where $(X, Y) \in \mathcal{X} \times \mathcal{Y}$ indicates a source data point with its label,
$k\in \{\textrm{II}, \textrm{III}\}$ is the level of the feature pyramid, and ${\mathcal{R}}_0^\textrm{II}$ and ${\mathcal{R}}_0^\textrm{III}$ are cross-relation features generated from $(X,Y)$;
$\theta \in \Theta$ is the set of weights of the entire model, 
$h$ is the classifier in terms of a fully-connected layer, 
$\ell$ is cross-entropy loss, $c$ is the transportation cost measured by the mean squared error and $\gamma$ is the penalty parameter of the transportation cost.
Notably, the spatiotemporal perturbations have a time dimension of $M$, which is equal to the number of segments of the input video, and thus enables the model to iteratively add the perturbations to adversarial examples.
After $T_\text{max}$ maximization phases, we convey the final $X^\text{II}$ and $X^\text{III}$ to the next minimization phase.

\subsubsection{Minimization Phase with Robust Training}

The next step is to optimize the classification loss over the augmented data on-the-fly. 
To further compensate for the potential effects caused by any overly divergent new data points and to prevent APN from being incorrectly generalized to similar but different categories, \textit{e.g.}, the misclassification of \textit{playing football} as \textit{playing basketball} as shown in Fig.~\ref{fig:intro_show}, we introduce a surrogate loss with robustness regularization over the hierarchical features learned by APN:
\begin{equation}
\label{cls-loss}
    L_{\textrm{cls}}^k(\theta ; ({X}^{k},Y)) := \underbrace{\ell(h(\mathcal{R}^k), Y)}_{ \text {classify adversarial examples}},
\end{equation}

\begin{equation}
\label{reg-loss}
     L_{\textrm{robust}}^k(\theta ; ({X}^{k},Y)) :=  \underbrace{\ell\left(h(\mathcal{R}^k), h\left(\mathcal{R}_{0}^\textrm{III}\right)\right)}_{\text {robustness regularization}},
\end{equation}

\begin{equation}
\label{robust-loss}
    L_{\textrm{min}} := L_{\textrm{cls}}^k(\theta ; ({X}^{k},Y)) + \lambda L_{\textrm{robust}}^k(\theta ; ({X}^{k},Y)),
\end{equation}
where $\lambda \ge 0$ is a hyperparameter for robustness regularization. As above, $\ell$ is cross-entropy loss, $h$ is the classifier in terms of a fully-connected layer, and $k \in \{\textrm{II}, \textrm{III}\}$ is the level of the feature pyramid.

\myparagraph{Why robust regularization? }In other words, we apply the objective function in Eq. \eqref{robust-loss} for both the cross-relation features at Level II and Level III of the pyramid. 
We believe that the generalizability of APN can be improved by enhancing the robustness of spatiotemporal features to the adversarial examples from different pyramid levels.
Such a regularization form was firstly introduced by Zhang \textit{et al.} \cite{ZhangYJXGJ19}. 
It is initially designed to strike a balance between accuracy and robustness in the design of defenses against adversarial examples.
Zhang \textit{et al.} showed that the robust error can in general be bounded tightly using two terms: one corresponds to the classification error measured by a surrogate loss function, and the other corresponds to how likely the input features are close to the extension of the decision boundary, termed as the boundary error.
The regularization term pushes the decision boundary away from the data points (Please refer to Fig. 1 in \cite{ZhangYJXGJ19}), making it more robust to perturbations. 
The video adversarial examples that are originated from different levels of APN are much more diverse than those of image data. We thus apply the robustness regularization term to trade source validation accuracy against generalization performance.
In other words, we use it to improve the ``smoothness'' of the categorical predictions generated by APN when using the source data and the corresponding augmented data.
In later experiments, we will find that when $\lambda$ increases, the source validation accuracy decreases while the generalization performance increases, in the sense that it controls a trade-off between in-domain accuracy and cross-domain robustness.   
Particularly, removing the robustness regularization term ($\lambda=0$) deteriorates generalization results to the greatest extent.
It is worth noting that we minimize the distance from both $h(\mathcal{R}^\textrm{III})$ and $h(\mathcal{R}^\textrm{II})$ to $h(\mathcal{R}_{0}^\textrm{III})$, because we want to further enhance the discriminability of APN. 
Compared with $\mathcal{R}^\textrm{II}_0$, the $\mathcal{R}^\textrm{III}_0$ feature is directly used to classify source data, and are therefore more discriminative. Minimizing the distance from $h(\mathcal{R}^\textrm{II})$ to $h(\mathcal{R}^\textrm{III}_0)$ can partly prevent $\mathcal{R}^\textrm{II}$ from being generalized to incorrect categories with similar temporal cues. The complete training framework is summarized in Alg. \ref{alg:Framwork}.

\section{Experiments}
\label{sec:expri}

In this section, we construct three video domain generalization benchmarks and use them to validate the effectiveness of VideoDG. 
On the UCF-HMDB benchmark, source and target domains are divided according to different datasets. 
On the multi-view NTU benchmark, domains are naturally divided according to different camera views. 
On the Something-Something benchmark, domains are divided according to different consequences of actions and video events such as \textit{doing something} vs. \textit{pretending to do something}.


\subsection{Experimental Setups}

\begin{figure*}[t]
	\centering
	\includegraphics[width=\textwidth]{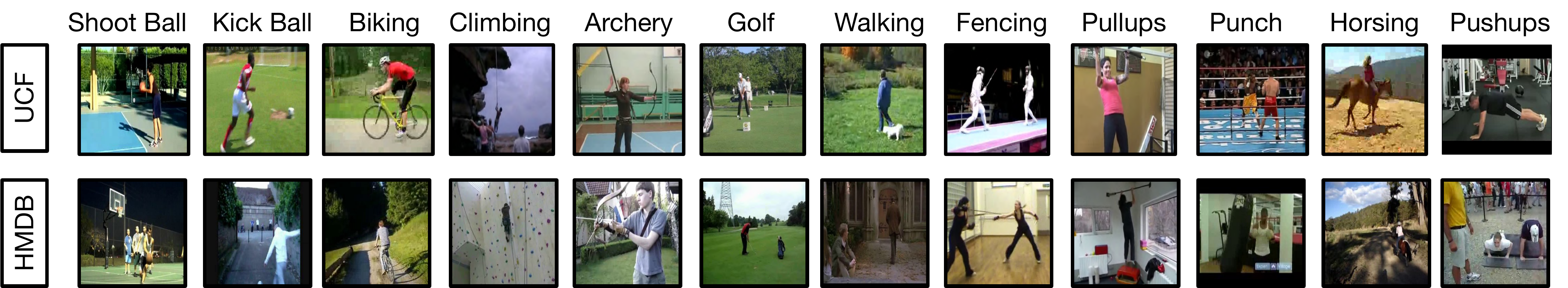}
	\vspace{-15pt}
	\caption{
		Snapshots of all 12 categories of the cross-dataset UCF-HMDB benchmark. Spatial and temporal domain shifts co-exist in these scenarios.
	}
	\label{fig:ucf-hmdb_show}
\end{figure*}

\begin{table*}[t]
\centering
\caption{Test accuracy ($\%$) on the UCF-HMDB benchmark: (1) APN is compared with existing video classification networks; (2) RADA is compared with existing domain generalization approaches. APN and RADA are shown to be firmly connected and improve each other. \revise{\textbf{VideoDG = APN + RADA.}} 
} 
\vspace{-5pt}

\resizebox{1\textwidth}{!}{
\setlength{\tabcolsep}{3pt}
    \begin{tabular}{lccccccc|ccccccc}
    \toprule
    \multirow{2}{0pt}{Model} 
    & \multicolumn{7}{c|}{UCF $\rightarrow$ HMDB} 
    & \multicolumn{7}{c}{HMDB $\rightarrow$ UCF} \\
    & Backbone & $\textrm{ADA}_\text{sem}$ & $\textrm{ADA}_\text{pixel}$& $\textrm{ADA}_\text{Alg. \ref{alg:baseline-Framwork}}$ & M-ADA & \textbf{RADA}  & Jigsaw 
    & Backbone & $\textrm{ADA}_\text{sem}$ & $\textrm{ADA}_\text{pixel}$ & $\textrm{ADA}_\text{Alg. \ref{alg:baseline-Framwork}}$ & M-ADA & \textbf{RADA}  & Jigsaw \\
    \midrule
    TSN & 51.4 $\pm$ 0.2 & 51.1 $\pm$ 0.3 &49.6 $\pm$ 0.3& 51.4 $\pm$ 0.2   & 52.4 $\pm$ 0.2 & 51.3 $\pm$ 0.2 &  51.5 $\pm$ 0.3   & 68.6 $\pm$ 0.3 & 68.2 $\pm$ 0.2 & 67.4 $\pm$ 0.2 & 68.3 $\pm$ 0.3   & 69.2 $\pm$ 0.2 & 68.3 $\pm$ 0.2  & 68.5 $\pm$ 0.3 \\ 
    TRN & 52.4 $\pm$ 0.3 & 52.8 $\pm$ 0.2 & 52.1 $\pm$ 0.3&52.9 $\pm$ 0.2   & 53.4 $\pm$ 0.3 & 53.9 $\pm$ 0.2  & 53.3 $\pm$ 0.3 &  69.8 $\pm$ 0.3 &69.6 $\pm$ 0.5 & 70.6 $\pm$ 0.2 & 70.0 $\pm$ 0.4   & 69.9 $\pm$ 0.3 & 71.2 $\pm$ 0.3  & 70.1 $\pm$ 0.3  \\
    TSM & 52.2 $\pm$ 0.3 & 51.3 $\pm$ 0.3 & 52.7 $\pm$ 0.3& 51.7 $\pm$ 0.3   & 52.5 $\pm$ 0.2  &51.3 $\pm$ 0.2  &52.5 $\pm$ 0.3 & 69.2 $\pm$ 0.3 & 68.6 $\pm$ 0.3 &  68.3 $\pm$ 0.2 &68.3 $\pm$ 0.3  & 69.1 $\pm$ 0.3  & 69.2 $\pm$ 0.2  & 68.9 $\pm$ 0.3  \\ 
    I3D &  52.2 $\pm$ 0.3 & 51.5 $\pm$ 0.3 & 51.5 $\pm$ 0.2 & 51.9 $\pm$ 0.2 & 52.1 $\pm$ 0.3 & 52.4 $\pm$ 0.3 &52.2 $\pm$ 0.3 &  68.6 $\pm$ 0.3 & 67.3 $\pm$ 0.3 & 67.1 $\pm$ 0.2 &67.8 $\pm$ 0.4  & 68.7 $\pm$ 0.3 & 68.5 $\pm$ 0.3  & 68.3 $\pm$ 0.2  \\
    NL-I3D & 54.0 $\pm$ 0.3 & 53.5 $\pm$ 0.4 &54.0 $\pm$ 0.2& 53.5 $\pm$ 0.3   & 54.9 $\pm$ 0.2 & 55.3 $\pm$ 0.3   &55.1 $\pm$ 0.4  & 71.2 $\pm$ 0.3 & 70.3 $\pm$ 0.3 & 71.1 $\pm$ 0.2&70.5 $\pm$ 0.3  & 71.9 $\pm$ 0.3 & 72.3 $\pm$ 0.3   & 72.0 $\pm$ 0.3 \\
    \textbf{APN} & 54.3 $\pm$ 0.3 & 55.2$\pm$ 0.3 & 56.9 $\pm$ 0.2& 55.5 $\pm$ 0.3  & 55.6 $\pm$ 0.3  & \textbf{59.1} $\pm$ 0.3  & 55.2 $\pm$ 0.4  & 71.4 $\pm$ 0.3& 71.9 $\pm$ 0.3 & 72.2 $\pm$ 0.3 &72.2 $\pm$ 0.2  & 71.5 $\pm$ 0.3 & \textbf{74.9} $\pm$ 0.3 & 72.4 $\pm$ 0.3 \\
    \bottomrule
    \end{tabular}
}
\label{tab:ucf_hmdb_left}
\end{table*}

\subsubsection{Source Validation Protocols}\label{validation}

\revise{Model selection protocols are crucial for all transfer learning approaches, especially when the test domains are not accessible during training. 
As described in the work from Gulrajani and Lopez-Paz \cite{gulrajani2020search}, existing domain generalization approaches often select the best-performing models and hyperparameters in the source validation set, but implicitly ignores its distribution shift from the test set. The hyperparameters that perform well on the validation set generally lack cross-domain generalizability.
}

\myparagraph{One-class SVM source validation.} 
\revise{We present an alternative scheme of in-domain model selection, without accessing the test domain. It can better ensure the cross-domain generalization ability of the selected models and hyperparameters.
Different from the regular training/validation splits, we explicitly mine and leverage the in-domain distribution shift to divide the source dataset. 
Specifically, for each video clip, we randomly sample $5$ frames and extract their deep features using a ResNet50 model \cite{HeRes} that is pretrained on ImageNet. We then average the values of these $5$ features and feed it to a one-class SVM \cite{scholkopf1999support} with the non-linear RBF kernel. We train the SVM over the entire source dataset, in which samples with lower scores are considered outliers. 
We take the top-scoring $70\%$ of the samples as the training set and the rest as the validation set. We tune hyperparameters and perform model selection on the new splits. In this way, the out-of-domain generalization ability of the models can be partially evaluated within the source domain. In Section \ref{sec:hyper}, we include the sensitivity analyses of hyperparameters.
}

\subsubsection{Implementation and Training Details} 

Due to the strong evidence that heavy data augmentation can improve domain generalization performance\cite{volpi2019addressing,jackson2019style}, we apply random cropping and horizontal flipping to the input frames of all compared models at training time.
%
%
%
%
%
%
We set the number of video segments $M$ to $5$ on all benchmarks, which is a common practice in the existing video classification and domain adaptation approaches \cite{Wang16,zhou2018temporal,chen2019temporal}. 
The random sampling strategy of input frames is determined by following that of previous models, \textit{e.g.}, TSN \cite{Wang16} and TRN \cite{zhou2018temporal}.
The number of feature levels used in RADA and the choice of cross-relation aggregation functions are determined in preliminary experiments by comparing with other options, \textit{e.g.}, element-wise sum and concatenation.

We run all experiments three times and report the mean and standard deviation of the results.
During the training procedure, we use the SGD optimizer with a batch size of $40$ video clips, and a base learning rate of $0.001$, which is reduced by $10$ times per $30$ epochs. 
Our models converge in $150$ epochs, taking less than $16$ hours on $8$ TITAN-X GPUs.


\subsubsection{Compared Models} 
\myparagraph{Video classification model.}
We compare APN with five widely-used or state-of-the-art video classification models: TSN \cite{Wang16}, TRN \cite{zhou2018temporal}, TSM \cite{lin2019tsm}, I3D \cite{carreira2017quo}, and NL-I3D \cite{NonLocal2018} (NL here stands for Non-Local). For a fair comparison, TSN, TSM, TRN, and our models all use the ResNet-50 \cite{HeRes} backbone that is pretrained on ImageNet. I3D and NL-I3D use the same 3D ResNet-50 architecture with $32$ input frames. These backbones are pretrained on the Kinetics dataset \cite{kay2017kinetics}. 
We also include baseline models by combining the above networks with the original image ADA method \cite{ADA}.

\myparagraph{Single-source domain generalization methods.}
Since our approach is an early work for video domain generalization, we compare it with baseline models that apply existing image-based methods for learning generalizable features to video data: 
\revise{
\begin{itemize}
    \item $\textrm{ADA}_\text{sem}$: The original ADA with an appended training set \cite{ADA}, which generates adversarial examples by maximizing the transportation cost in the semantic space as shown in Eq.~\eqref{max-phase}.
    \item $\textrm{ADA}_\text{pixel}$: We re-design the transportation cost by defining the distance in the pixel space. For both $\textrm{ADA}_\text{sem}$ and $\textrm{ADA}_\text{pixel}$, we set $T_\textrm{max}=10$ (\textit{i.e.}, the number of maximization phases) and $K=3$ (\textit{i.e.}, the number of minimax training procedures to append the source dataset).
    \item $\textrm{ADA}_\text{Alg. \ref{alg:baseline-Framwork}}$: It can be viewed as a memory-efficient version of $\textrm{ADA}_\text{sem}$ that is optimized over the adversarial examples on-the-fly instead of using them to append the training set. We here set $T_\textrm{max}=5$ to the same value as RADA.
    \item We also apply M-ADA \cite{qiao2020learning}, Jigsaw puzzle \cite{Jigsaw}, and data augmentation methods (\textit{e.g.}, image rotation and color jittering) to the same network backbone of our approach. For M-ADA, we set $K=4$.
\end{itemize}
}

\myparagraph{Multi-source domain generalization methods.}
Finally, we include existing image-based methods that can learn generalizable features from multiple source domains such as Epi-FCR \cite{li2019episodic}, MMD-AAE \cite{li2018DG}, and CIDDG \cite{li2018deep}. 

All the compared models, including VideoDG, use the same ensemble strategy at test time with different values of $\gamma \in \{0.001, 0.01, 0.1, 1, 10\}$.

We show that the new video domain generalization problem is so challenging that its performance cannot be greatly improved by simple combinations of existing image-based domain generalization methods and previous video classification models.

\begin{table}[t]
	\centering
	\small
	\caption{
	Results ($\%$) of applying different data augmentation techniques to APN.
	All models are trained with image cropping and horizontal flipping.
	\revise{\textit{Rand aug.} \cite{volpi2019addressing} combines all other augmentation techniques except for RADA.} 
	} 
	\vspace{-5pt}
	\begin{tabular}{lccc}
		\toprule
		Method & UCF $\rightarrow$ HMDB & HMDB $\rightarrow$ UCF \\
		\midrule
		APN backbone & 54.3 $\pm$ 0.3 & 71.4 $\pm$ 0.3 \\
		Image rotation & 55.4 $\pm$ 0.2 & 72.1 $\pm$ 0.2  \\ 
		Color jittering & 55.0 $\pm$ 0.3 & 71.9 $\pm$ 0.3 \\
		Contrast perturb. &  54.9 $\pm$ 0.3 & 71.7 $\pm$ 0.3  \\
		Brightness perturb. & 55.2 $\pm$ 0.3 & 71.5 $\pm$ 0.3  \\
		\revise{Rand aug.} & 56.1 $\pm$ 0.2  &  72.6 $\pm$ 0.3\\
		VideoDG  & \textbf{59.1} $\pm$ 0.3 & \textbf{74.9} $\pm$ 0.3  \\
		\bottomrule
	\end{tabular}
	\label{tab:ucf_hmdb_data_augmentation}
\end{table}

\begin{figure}[t]
    \centering
    \includegraphics[width=\columnwidth]{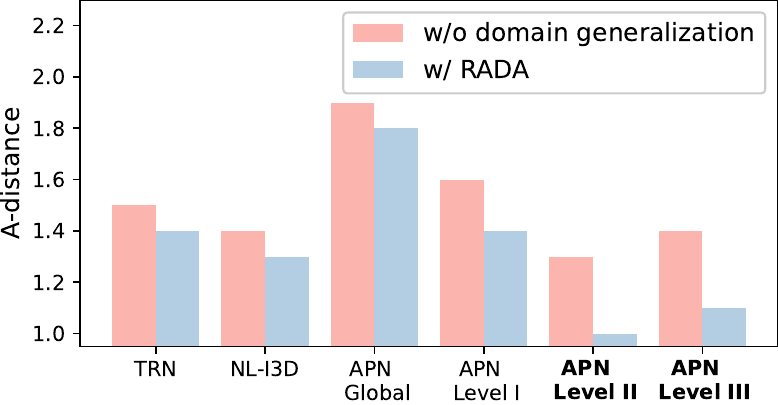}
    \vspace{-10pt}
    \caption{A-distances of different models/features that measure the domain shift from UCF to HMDB. A lower A-distance indicates a model that can better generalize across domains. 
    }
    \label{fig:A-distance}
    \vspace{-5pt}
\end{figure}

\begin{table}[t]
\centering
\small
\caption{The notation of number of samples that were (in)correctly classified before the use of RADA, and are (in)correctly classified after that. 
} 
\vspace{-5pt}
\renewcommand{\multirowsetup}{\centering}  
    \begin{tabular}{ll|cc}
    \toprule
    & & \multicolumn{2}{c}{Before the use of RADA} \\
    & & True & False\\
    \midrule
    \multirow{2}{1cm}{After} & Positive & TP & FP \\
    & Negative & TN & FN \\
    \bottomrule
    \end{tabular}
\label{tab:correction_rate}
\end{table}

\begin{figure}[t]
    \centering
    \includegraphics[width=1\columnwidth]{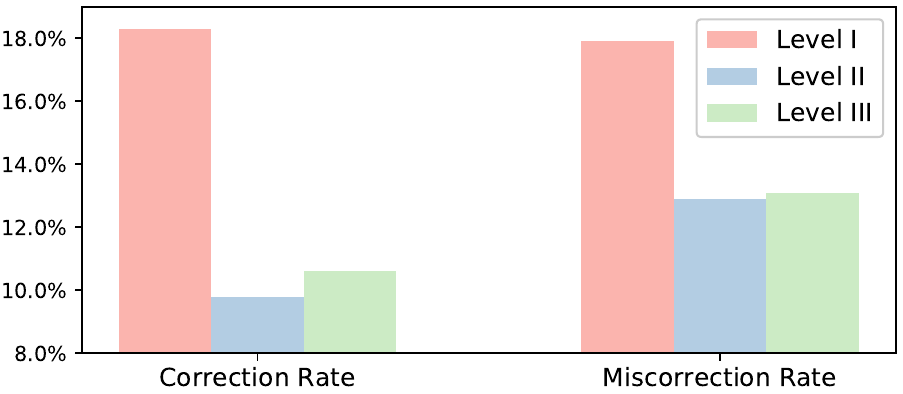}
    \vspace{-10pt}
    \caption{We use the correction rate (CR) and the miscorrection rate (MR) to evaluate the roles of features at different levels on UCF $\rightarrow$ HMDB.
    }
    \label{fig:correction_rate}
    \vspace{-5pt}
\end{figure}

\begin{figure*}[t]
    \centering
    \includegraphics[width=0.95\textwidth]{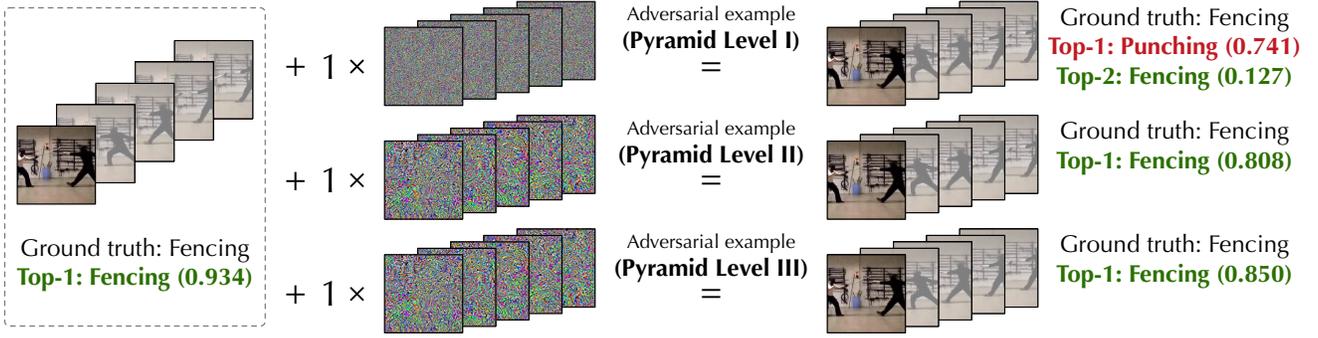}
    \vspace{-8pt}
    \caption{The visualization of adversarial examples at different pyramid levels and corresponding classification results of VideoDG.
    There are subtle differences between the perturbed frames at Level I and those at Levels II/III.
    }
    \label{fig:adv-showcase}
    \vspace{-5pt}
\end{figure*}

\begin{table}[t]
\centering
\small
\caption{Test accuracy ($\%$) on UCF-HMDB using APN as the network backbone. Indexes on RADA indicate the feature levels used to generate new data.
}
\addtolength{\tabcolsep}{-5pt}
\vspace{-5pt}
    \begin{tabular}{lccc}
    \toprule
    Model & UCF $\rightarrow$ HMDB & HMDB $\rightarrow$ UCF \\
    \midrule
    $\textrm{RADA}_{\textrm{I}}$ & 53.6 $\pm$ 0.3 & 70.5 $\pm$ 0.3  \\ 
    $\textrm{RADA}_{\textrm{II}}$ & 57.1 $\pm$ 0.3 & 73.6 $\pm$ 0.3 \\
    $\textrm{RADA}_{\textrm{III}}$ & 56.4 $\pm$ 0.4 & 73.1 $\pm$ 0.3  \\ 
    $\textrm{RADA}_{\textrm{I},\textrm{II}}$ & 56.9 $\pm$ 0.3 & 72.9 $\pm$ 0.4\\
    $\textrm{RADA}_{\textrm{I},\textrm{III}}$ & 56.1 $\pm$ 0.4 & 72.7 $\pm$ 0.3 \\
    $\textrm{RADA}_{\textrm{II}, \textrm{III}}$ (Final) & \textbf{59.1} $\pm$ 0.3 & \textbf{74.9} $\pm$ 0.3  \\
    $\textrm{RADA}_{\textrm{I},\textrm{II}, \textrm{III}}$ & 58.6 $\pm$ 0.3 & 74.3 $\pm$ 0.3  \\
    \bottomrule
    \end{tabular}
\label{tab:ucf_hmdb_right}
\vspace{-5pt}
\end{table}

\subsection{UCF-HMDB: A Cross-Dataset Benchmark}

We first evaluate our approach through generalization experiments across datasets. The UCF-HMDB benchmark consists of the $12$ overlapping categories shared by UCF101 \cite{UCF101} and HMDB51 \cite{HMDB51}, which contains a total of $3{,}809$ videos. 
The benchmark is adopted from existing work for video domain adaptation \cite{chen2019temporal}.
We take one dataset as the source domain and the other as the target one. 
In Fig. \ref{fig:ucf-hmdb_show}, we show frame examples for each category. Videos from different domains (datasets) may have diverse action behaviours, \textit{e.g.}, practicing shooting or playing the whole basketball game.
%
%
The domain shift also exists in the spatial information, \textit{e.g.}, playing football on the field or in an alley, climbing outdoors or indoors.

\subsubsection{The Challenging Video Domain Generalization} 

As shown in Table \ref{tab:ucf_hmdb_left}, most compared models, including TSN, TSM, I3D, and NL-I3D, show performance degradation when using the original ADA method. For example, the performance of the I3D-based Non-Local model reduces from $54.0\%$ to $53.5\%$ after applying ADA. These results may look counter-intuitive, but indicate that video domain generalization is a super challenging task that cannot be greatly improved by directly applying existing methods of learning generalizable image features to the video classification backbones. An effective domain augmentation method for spatiotemporal data requires unique considerations for the temporal domain shift.

\subsubsection{Main Results}

\myparagraph{Closely related APN and RADA.}
As shown in Table \ref{tab:ucf_hmdb_left}, VideoDG (APN with RADA) achieves the best test accuracy of $59.1\%$ (UCF $\rightarrow$ HMDB).
Compared with other video classification networks, it largely exceeds the second place, \textit{i.e.}, NL-I3D with RADA, by $3.8\%$.
Besides, we have two interesting observations from Table \ref{tab:ucf_hmdb_left}. First, the APN model consistently works well with all the compared domain generalization approaches, indicating that its relational pyramid provides more generalizable features. 
Second, the RADA algorithm particularly performs better on APN than on any other backbones. It is because RADA relies on the relational features of multiple pyramid levels to compromise between the diversity and the focus of the generated adversarial examples. In this sense, the RADA algorithm is intimately coupled with the proposed APN architecture.

\myparagraph{Comparison with other ADA-based methods.}
We compare RADA with other domain generalization methods that are based on adversarial perturbations and modified for video data.
From Table \ref{tab:ucf_hmdb_left}, RADA significantly outperforms the original ADA with a progressively appended training set, the memory-efficient ADA shown in Alg. \ref{alg:baseline-Framwork}, and M-ADA which is the state-of-the-art method for single-domain image domain generalization. All the above approaches are based on the APN model.
Furthermore, RADA also outperforms a variant of $\textrm{ADA}_\text{pixel}$  that is also modified for video inputs by $2.2\%$ from UCF to HMDB and $2.7\%$ backward. The only difference between them is the form of transportation cost in \eqn{max-phase}, where RADA uses the semantic space constraint while $\textrm{ADA}_\text{pixel}$ uses the pixel space constraint based on the Wasserstein distance.

\myparagraph{Comparison with traditional (hard) data augmentation.}
Inspired by the previous study that traditional (hard) data augmentation techniques can improve the generalization ability of image classification models \cite{volpi2019addressing}, we here explore the benefits of image rotation, color jittering, contrast perturbation, and brightness perturbation in the video scenarios.
\revise{We also adopt the training scheme of randomized data augmentation from Volpi and Murino \cite{volpi2019addressing}, and sequentially apply all the transformations to 
each minibatch of source data.}
As shown in Table \ref{tab:ucf_hmdb_data_augmentation}, the RADA algorithm improves the APN backbone by a larger margin than other data augmentation methods.
We argue that, in video scenarios, the spatial transformations cannot greatly benefit the generalization performance, 
without augmenting the training set from the perspective of temporal relations.

\myparagraph{Comparison with the Jigsaw puzzle method.}
Solving the Jigsaw puzzle has been shown effective for improving the generalizability of image classification models \cite{Jigsaw}, so we include this method in the experiments.
We here refer to \cite{ahsan2019video} to generate Jigsaw puzzles from the video frame patches.
More specifically, in a tuple of $M$ sampled frames, each video frame is divided into $4$ patches, and we use all the $4 \times M$ frame patches to construct the Jigsaw puzzle. 
As shown in Table \ref{tab:ucf_hmdb_left}, the modified Jigsaw puzzle method obtains a similar performance to the original ADA method, but its results are still inferior to the proposed RADA algorithm.

\subsubsection{Generalizability Analysis by A-distance}
\label{sec:a-distance}
As mentioned above, we use the A-distance \cite{Ben-DavidBCKPV10} to quantify the generalizability (how well it can mitigate the domain shift). It is defined as
\begin{equation}
    d_{A}=2(1-2 \epsilon),
\end{equation}
where $\epsilon$ is the error rate of a domain classifier trained to discriminate the source domain and the target domain.

Fig.~\ref{fig:A-distance} compares the A-distances of different levels of APN features from the UCF domain to the HMDB domain.
We can see that the global-relation features produce the highest A-distance, being less generalizable than any representations of APN.
It also shows that the A-distances of Level-II and Level-III features (cross-relation) are much smaller than that of the Level-I features (local-relation), leading us to design the RADA framework based on cross-relation features to adjust the focus of the expanded source domain.

Further, compared with NL-I3D, we can see that APN can better reduce the temporal domain shift, though it has fewer parameters. More importantly, the even smaller A-distances after applying RADA show that the effectiveness comes from its ability to reduce domain shift rather than solely relying on APN as a more powerful deep architecture.

\begin{table}[t]
	\centering
	\small
	\caption{
	Test accuracy ($\%$) of using different numbers of maximization phases ($T_\textrm{max}$). All models are based on the APN backbone.
	} 
	\vspace{-5pt}
	\begin{tabular}{ccc|cc}
		\toprule
		\multirow{2}{4pt}{$T_\textrm{max}$} 
		& \multicolumn{2}{c|}{UCF $\rightarrow$ HMDB} 
		& \multicolumn{2}{c}{HMDB $\rightarrow$ UCF} \\
		  & RADA & $\textrm{ADA}_\text{pixel}$  & RADA & $\textrm{ADA}_\text{pixel}$ \\
		\midrule
		 1 & 58.1 $\pm$ 0.3 & 55.8 $\pm$ 0.3  & 74.1 $\pm$ 0.3 & 70.9 $\pm$ 0.2  \\ 
		 5& \textbf{59.1} $\pm$ 0.3 & 56.5 $\pm$ 0.3 & \textbf{74.9} $\pm$ 0.3 & 71.7 $\pm$ 0.4 \\
		 10 &  58.8 $\pm$ 0.3 & \textbf{56.8} $\pm$ 0.4  &74.8 $\pm$ 0.3 & \textbf{72.1} $\pm$ 0.3 \\
		 15 & 58.1 $\pm$ 0.2 & 56.6 $\pm$ 0.3 & 74.5 $\pm$ 0.3 & 71.8 $\pm$ 0.3   \\
		\bottomrule
	\end{tabular}
	\label{tab:ucf_hmdb_Tmax}
	\vspace{-5pt}
\end{table}

\begin{figure*}[t]
  \centering
  \subfigure[Train/Val.: UCF, Test: HMDB]{
    \includegraphics[width=\columnwidth]{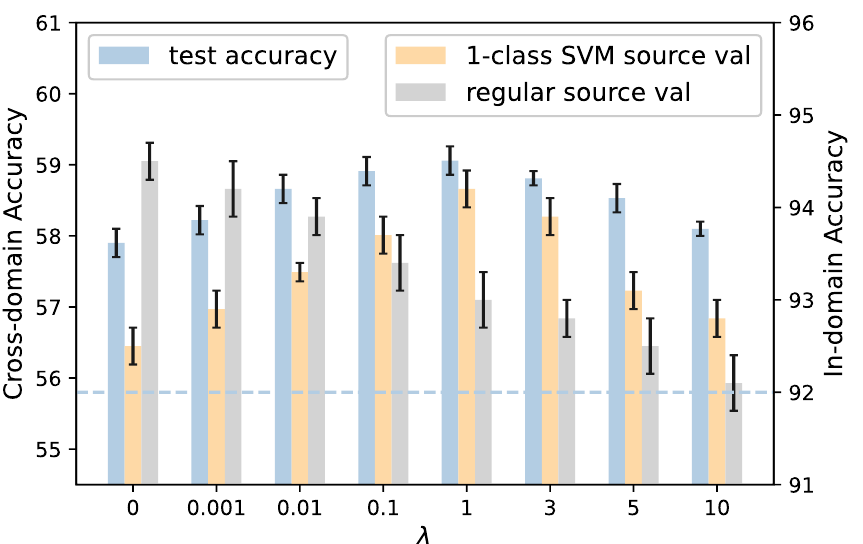}}
   \subfigure[Train/Val.: HMDB, Test: UCF]{
     \includegraphics[width=\columnwidth]{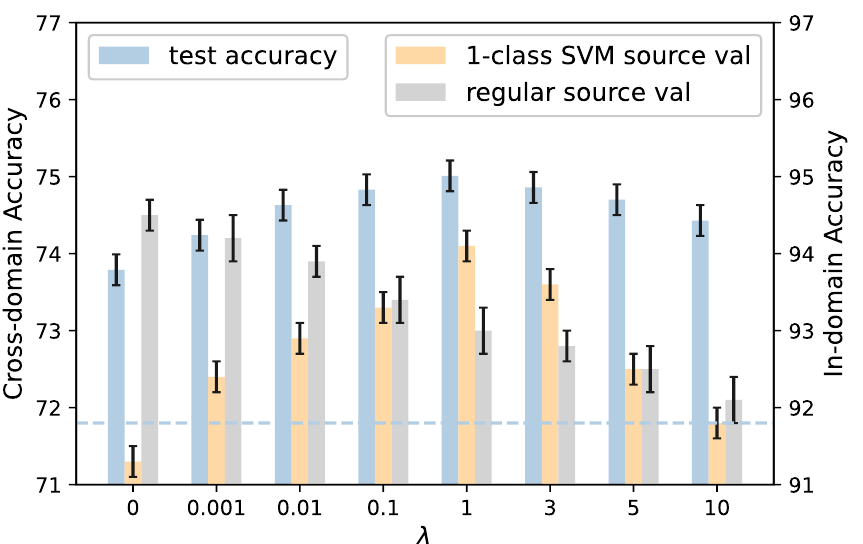}}
  \subfigure[Train/Val.: UCF, Test: HMDB]{
    \includegraphics[height=0.62\columnwidth]{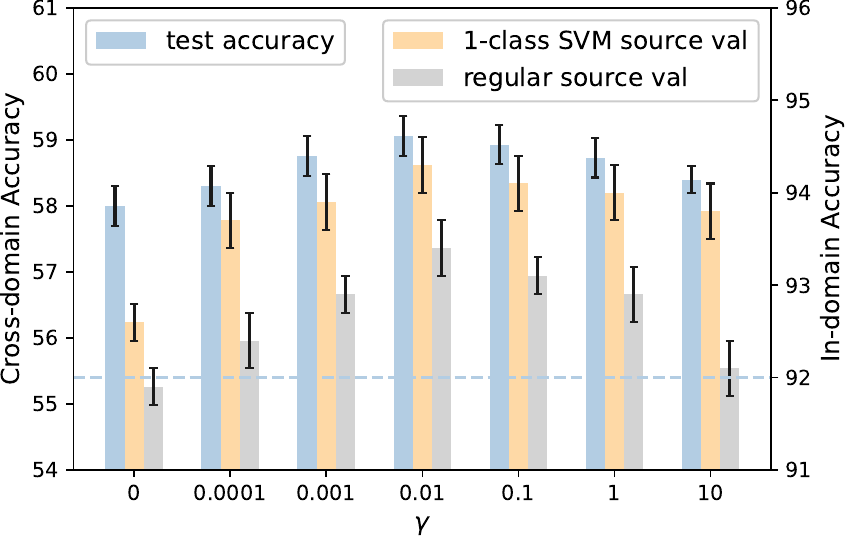}}
  \subfigure[Train/Val.: HMDB, Test: UCF]{
     \includegraphics[height=0.62\columnwidth]{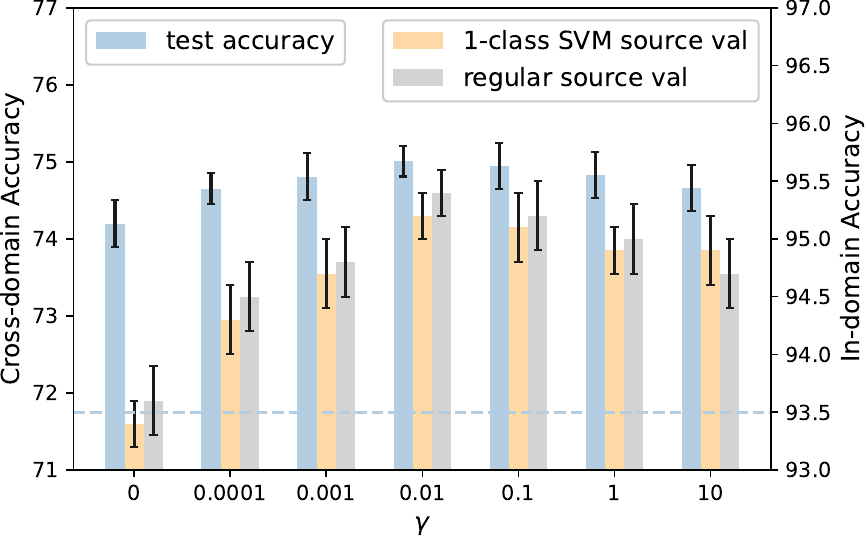}}
    \vspace{-5pt}
  \caption{The sensitivity analyses of the hyperparameters of VideoDG. 
  \revise{To show that the one-class SVM source validation protocol can facilitate hyperparameter tuning, we report both the in-domain performance using the regular and new training/validation splits.}
  For (a) and (b), we fix $\gamma=0.01$ and tune $\lambda$.
  \revise{
  The blue bars show the performance on the test set under the one-class SVM source validation protocol. On the regular splits, we can see from the gray bars that when $\lambda$ increases, the source validation accuracy decreases while the generalization performance increases. On the one-class SVM splits, as shown by the orange bars, the best-performing model generalizes well to the test domain.} For (c) and (d), we fix $\lambda=1$ and tune $\gamma$. \revise{The hyperparameters are robust such that all models consistently outperform $\textrm{ADA}_\text{sem}$ (dashed blue line) on the test set.}
  }
  \label{fig:hyper}
\end{figure*}

\begin{table*}[b]
\centering
\small
\caption{Test accuracy ($\%$) on the NTU benchmark with multiple target domains. \textbf{L} (left), \textbf{C} (center), and \textbf{R} (right) are different viewpoint domains. 
} 
\vspace{-10pt}
    \begin{tabular}{lcc|cc|cc}
    \toprule
    \multirow{2}{0pt}{Model} 
    & \multicolumn{2}{c|}{ L $\rightarrow$ C, R} 
    & \multicolumn{2}{c|}{C $\rightarrow$ L, R}
    & \multicolumn{2}{c}{R $\rightarrow$ L, C}\\
    & Backbone & RADA 
    & Backbone & RADA  
    & Backbone & RADA  \\
    \midrule
  TSN   & 39.5 $\pm$ 0.3 & 39.2 $\pm$ 0.3  &  60.6 $\pm$ 0.3 & 58.9 $\pm$ 0.3  & 47.5 $\pm$ 0.2 & 46.7 $\pm$ 0.4 \\ 
    TRN &  43.6 $\pm$ 0.2 & 43.9 $\pm$ 0.3  & 62.2 $\pm$ 0.3 & 62.6 $\pm$ 0.3  & 49.8 $\pm$ 0.3 & 49.9 $\pm$ 0.2  \\
    TSM & 40.7 $\pm$ 0.2 & 39.3 $\pm$ 0.3  &  60.8 $\pm$ 0.2 & 59.7 $\pm$ 0.3  & 48.7 $\pm$ 0.2 & 47.6 $\pm$ 0.3 \\
    I3D &   41.4 $\pm$ 0.3 & 41.3 $\pm$ 0.3    & 51.8 $\pm$ 0.1 & 50.3 $\pm$ 0.2  & 42.0 $\pm$ 0.3 & 41.9 $\pm$ 0.2  \\ 
    NL-I3D &  42.2 $\pm$ 0.3 & 43.4 $\pm$ 0.4  & 52.9 $\pm$ 0.2 & 53.5 $\pm$ 0.2  & 43.5 $\pm$ 0.3 & 43.6 $\pm$ 0.3 \\
    APN &  44.9 $\pm$ 0.2 & \textbf{49.4} $\pm$ 0.2 & 63.7 $\pm$ 0.2 & \textbf{68.6} $\pm$ 0.2 & 53.1 $\pm$ 0.3 & \textbf{55.6} $\pm$ 0.2   \\
    \bottomrule
    \end{tabular}
\label{tab:ntu}
\vspace{-10pt}
\end{table*}

\subsubsection{Different Roles of Pyramid Levels in RADA}

First and foremost, as shown by the A-distances in Fig.~\ref{fig:A-distance}, the Level-I local relational features are more generalizable than the global relational features. 

Next, we evaluate the ratio of test samples misclassified by the APN without RADA but corrected by applying RADA at each pyramid levels, namely the Correction Rate: $\textrm{CR} = \textrm{FP}/(\textrm{FP}+\textrm{FN})$, where the definition of $\textrm{FP}$ and $\textrm{FN}$ is shown in Table \ref{tab:correction_rate}.
We can see from Fig.~\ref{fig:correction_rate} that $\textrm{CR}_{\text{Level-}\textrm{I}}$ (18.3\%) $>$ $\textrm{CR}_{\text{Level-}\textrm{III}}$ (10.6\%) $>$ $\textrm{CR}_{\text{Level-}\textrm{II}}$ (9.8\%). 
On the other hand, we also evaluate the ratio of samples that were classified correctly before applying RADA but incorrectly after that, which is regarded as the Miscorrection Rate: $\textrm{MR} = \textrm{TN}/(\textrm{TP}+\textrm{TN})$,
%
where the definition of $\textrm{TP}$ and $\textrm{TN}$ is also shown in Table \ref{tab:correction_rate}.
From Fig.~\ref{fig:correction_rate}, we have 
$\textrm{MR}_{\text{Level-}\textrm{I}}$ (17.9\%) $>$ $\textrm{MR}_{\text{Level-}\textrm{III}}$ (13.1\%) $>$ $\textrm{MR}_{\text{Level-}\textrm{II}}$ (12.9\%). 
The above results indicate that the Level-I relational features enable the model to handle difficult samples, but can be more likely to trigger negative generalization effects at the same time. In contrast, the Level-II and Level-III relational features improve the generalizability in a more conservative manner.

Fig.~\ref{fig:adv-showcase} further illustrates the need to use cross-relation features with a visual example. A video of \textit{fencing} is misclassified as \textit{punching} by the APN trained with Level-I adversarial examples. The incorrect generalization can be caused by similar short-term motion that is less discriminative from the classification view.

We also observe that of all the samples correctly classified by applying APN at Level II ( $\textrm{RADA}_{\textrm{II}}$), only $38.3\%$ are also correctly classified by applying APN at Level III ($\textrm{RADA}_{\textrm{III}}$). It gives strong evidence that Level-II and -III features play complementary roles.

We summarize the above empirical results. First, the global relational features have the largest A-distance (the lower, the better). Second, the Level-I local relational features have the largest miscorrection rate (also the lower, the better). Third, the two levels of cross-relation features are complementary in the generalization results. 
These observations inspire us to use both Level-II and Level-III features of APN to generate adversarial examples in RADA. As shown in Table \ref{tab:ucf_hmdb_right}, it achieves the highest classification accuracy across domains, compared with other baselines.

\subsubsection{Hyperparameter Analyses} 
\label{sec:hyper}

\revise{
All hyperparameters are tuned on the source validation set, which is constructed by mining the in-domain distribution shift via a one-class SVM described in Section~\ref{validation}. 
}

\myparagraph{The number of maximization phases.}
Table \ref{tab:ucf_hmdb_Tmax} shows the test accuracy of RADA and $\textrm{ADA}_\text{pixel}$ with $T_\textrm{max} \in \{1, 5, 10, 15\}$, respectively. 
From the results, we can see that RADA consistently outperforms $\textrm{ADA}_\text{pixel}$ with different numbers of maximization phases. 
Besides, RADA yields the best results when $T_\textrm{max}=5$ and appears to be stable in the near range. We set the same $T_\textrm{max}$ value on other benchmarks as well.

\begin{figure*}[t]
    \centering
    \includegraphics[width=0.95\textwidth]{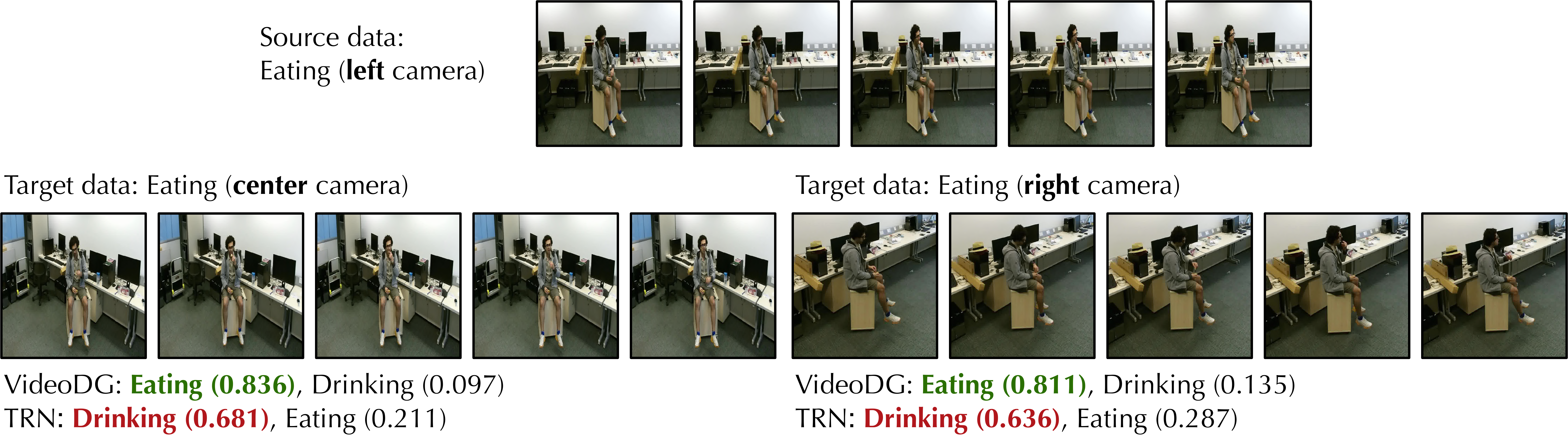}
    \vspace{-10pt}
    \caption{
    Two generalization examples on the multi-view NTU benchmark. The first row shows training data from the source domain. The second row shows the test data from the target domains. The green indicate making correct predictions and the red incorrect ones. 
    }
    \label{fig:ntu_show}
    \vspace{-5pt}
\end{figure*}

\begin{table}[t]
\centering
\small
\caption{
Test accuracy ($\%$) on NTU with multiple source domains. VideoDG (APN with RADA) is compared with existing image-based methods particularly designed for multi-source domain generalization.
}
\vspace{-10pt}
    \begin{tabular}{lccc}
    \toprule
    Model  & L, R $\rightarrow$ C & L, C $\rightarrow$ R & C, R $\rightarrow$ L \\
    \midrule
    Epi-FCR & 71.0 $\pm$ 0.2 &  58.5 $\pm$ 0.2 & 67.2 $\pm$ 0.3\\
    MMD-AAE  & 69.9 $\pm$ 0.3 & 57.3 $\pm$ 0.3 & 66.4 $\pm$ 0.2\\
    CIDDG  & 69.5 $\pm$ 0.3 & 56.0 $\pm$ 0.3 &  64.0 $\pm$ 0.2\\
    VideoDG & \textbf{76.1} $\pm$ 0.2 & \textbf{63.2} $\pm$ 0.2 & \textbf{72.5} $\pm$ 0.3\\
    \bottomrule
    \end{tabular}
\label{tab:ntu_DG}
\vspace{-5pt}
\end{table}

\myparagraph{Hyperparameter tuning for robustness regularization.}
\revise{
Under the one-class SVM source validation protocol, as shown by the orange bars in Fig.~\ref{fig:hyper} (a-b), the proposed model achieves the best validation accuracy when $\lambda=1$, which is consistent with the test domain performance.
By contrast, as shown by the gray bars, the model performs best on regular validation splits when $\lambda=0$. All in all, by using the one-class SVM validation strategy, we can select more generalizable models without accessing the test domain.}
Furthermore, since RADA has two contributions on top of ADA, one might wonder which will give more performance gains: the multi-scale adversarial examples that are originated from different feature levels of APN; or the robustness regularization.
To find the answer, we first disable the robustness loss in \eqn{robust-loss} by setting $\lambda=0$. As shown by the blue bars in Fig.~\ref{fig:hyper} (a-b), the test accuracy is slightly lower than that of $\lambda=1$. However, it is still higher than that of $\textrm{ADA}_\text{sem}$, as well as the memory-efficient $\textrm{ADA}_\text{Alg. \ref{alg:baseline-Framwork}}$. The results show that both contributions have important effects on the performance gain.

\myparagraph{Model ensembling with different hyperparameters for the transportation cost.}
\revise{
As shown in Fig.~\ref{fig:hyper} (c-d), we can easily tune the hyperparameter of $\gamma$ on the one-class SVM validation set. Different from the selection process of $\lambda$, at test time, we follow the original ADA method to ensemble $5$ separate models with different values of $\gamma \in \{0.001, 0.01, 0.1, 1, 10\}$, since these models have relatively small performance gaps on the validation set.
}


\begin{figure}[t]
    \centering
    \includegraphics[width=0.95\columnwidth]{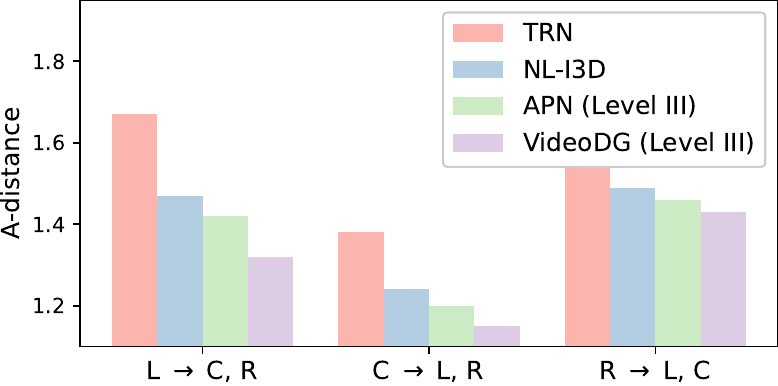}
    \vspace{-10pt}
    \caption{
    We calculate the A-distance of the Level-III features of APN (w/o or w/ RADA, as shown by the green and purple bars, respectively) and compare it with those of TRN and NL-I3D on the NTU benchmark.
    }
    \label{fig:ntu_a_distance}
    \vspace{-5pt}
\end{figure}

\subsection{NTU: A Cross-View Video Benchmark} 

It is a natural application scenario of video domain generalization to directly deploy a model trained with data recorded from one camera view in a new environment with videos from other camera views.
Here, we use the multi-view NTU \cite{Shahroudy_2016_CVPR} dataset, which contains around $57{,}000$ videos and $60$ categories of human activities. We build the benchmark by splitting the source and target domains according to three major viewpoints.
As different domains share the same temporal events, we can separately analyze the case of spatial domain shift on this benchmark.

\subsubsection{Main Results}

As shown in Table \ref{tab:ntu} and Table \ref{tab:ntu_DG}, VideoDG (\textit{i.e.}, APN trained with RADA) outperforms other models significantly under experimental setups with multiple target domains or multiple source domains.
The example in Fig.~\ref{fig:ntu_show} shows the spatial domain shift across different camera views. 
We can conclude that although our approach is specifically designed to deal with problems caused by the absence and misalignment of video events in different domains, it remains effective to overcome the general spatiotemporal domain shift, even in extreme cases where only spatial domain shift exists.
This conclusion is further supported by the results of A-distance in Fig.~\ref{fig:ntu_a_distance}. 
Compared with TRN, which is also a deep network based on temporal relational features, the proposed VideoDG reduces the domain shift more effectively on the NTU benchmark.

\begin{table}[t]
	\centering
	\small
	\caption{
	Test accuracy ($\%$) of VideoDG (APN with RADA) on the NTU benchmark using different numbers of video segments.
	}
	\vspace{-5pt}
	\label{tab:ntu_M}
	\begin{tabular}{lccc}
		\toprule
		$M$  & L, R $\rightarrow$ C & L, C $\rightarrow$ R & C, R $\rightarrow$ L \\
		\midrule
		3 & 44.3 $\pm$ 0.2 &  59.5 $\pm$ 0.2 & 50.2 $\pm$ 0.2 \\
		4   & 46.9 $\pm$ 0.3 & 62.5 $\pm$ 0.3 & 52.3 $\pm$ 0.2 \\
		5 & \textbf{49.4} $\pm$ 0.2 & 68.6 $\pm$ 0.2 & \textbf{55.6} $\pm$ 0.2\\
		6 & 49.2 $\pm$ 0.2 & \textbf{69.1} $\pm$ 0.2 & 55.2 $\pm$ 0.2 \\
		7 & 49.3 $\pm$ 0.3 & 68.9 $\pm$ 0.2 & 55.5 $\pm$ 0.3 \\
		\bottomrule
	\end{tabular}
	\vspace{-5pt}
\end{table}

\subsubsection{The Number of Video Segments}
In Table \ref{tab:ntu_M}, we analyze the effect of the number of video segments in the range of $M\in \{3,4,5,6,7\}$, and observe that, in most cases, VideoDG (APN with RADA) performs best when $M=5$. 
We also observe that the performance is quite stable when $M\geq 5$. 
For a trade-off between accuracy and efficiency, we choose $M=5$ for all benchmarks, which is also a typical setting of TRN \cite{zhou2018temporal}.

\subsubsection{Comparisons with Existing Multi-Source Domain Generalization Methods}

Besides the above benchmarks with a single source domain, on the multi-view NTU benchmark, we also follow the multi-source domain generalization setups to compare our approach with the previous methods, including Epi-FCR \cite{li2019episodic}, CIDDG \cite{li2018deep} and MMD-AAE \cite{li2018DG}. 
We use two views as source domains and the rest as the target one. As shown in Table \ref{tab:ntu_DG}, the proposed VideoDG approach remarkably outperforms other baselines. 
All compared models have very similar network backbones and similar numbers of parameters.
The results show that VideoDG not only performs well under the single-domain generalization but also achieves the best performance under the multi-source generalization setups.

\begin{table}[t]
\centering
\small
\caption{Test accuracy ($\%$) on the Something-Something benchmark.
} 
\vspace{-5pt}
    \begin{tabular}{lcc|cc}
    \toprule
    \multirow{2}{0pt}{Model} 
    & \multicolumn{2}{c|}{Source $\rightarrow$ Target} 
    & \multicolumn{2}{c}{Target $\rightarrow$ Source } \\
    & Backbone & RADA 
    & Backbone & RADA  \\
    \midrule
   TSN & 35.1 $\pm$ 0.3 & 35.2 $\pm$ 0.2  &  22.7 $\pm$ 0.2 & 22.3 $\pm$ 0.3   \\ 
		TRN & 36.8 $\pm$ 0.3 & 37.1 $\pm$ 0.1 & 23.5 $\pm$ 0.2 & 23.9 $\pm$ 0.3 \\
		TSM & 36.5 $\pm$ 0.2 & 36.3 $\pm$ 0.2   & 23.0 $\pm$ 0.3 & 23.0 $\pm$ 0.2 \\
		I3D &  31.0 $\pm$ 0.2 & 30.3 $\pm$ 0.2  & 22.7 $\pm$ 0.2 & 21.4 $\pm$ 0.3  \\ 
		NL-I3D & 33.2 $\pm$ 0.3 & 33.3 $\pm$ 0.2 & 23.2 $\pm$ 0.4 & 23.4 $\pm$ 0.2 \\
		APN & 38.0 $\pm$ 0.3&  \textbf{41.2} $\pm$ 0.2  & 24.5 $\pm$ 0.3 & \textbf{27.9} $\pm$ 0.2  \\
    \bottomrule
    \end{tabular}
\label{tab:sth}
\end{table}

\subsection{Something-Something: A Cross-Relation Video Domain Generalization Benchmark}  

In many practical applications, videos at test time may have similar but different consequences of actions, or different combinations of local temporal relations from the training source domain. Under these circumstances, the temporal domain shift is significant. 
We construct a benchmark by selecting $20$ basic categories from the Something-Something dataset \cite{something-something}. Under each category such as \textit{tearing}, there are two sub-categories with different consequences of actions such as \textit{tearing something} and \textit{pretending to be tearing something that is not tearable}. We assign them to different domains.
The source domain has $9{,}530$ videos and the target one has around $4{,}000$ videos.

\begin{figure}[t]
    \centering
    \includegraphics[width=\columnwidth]{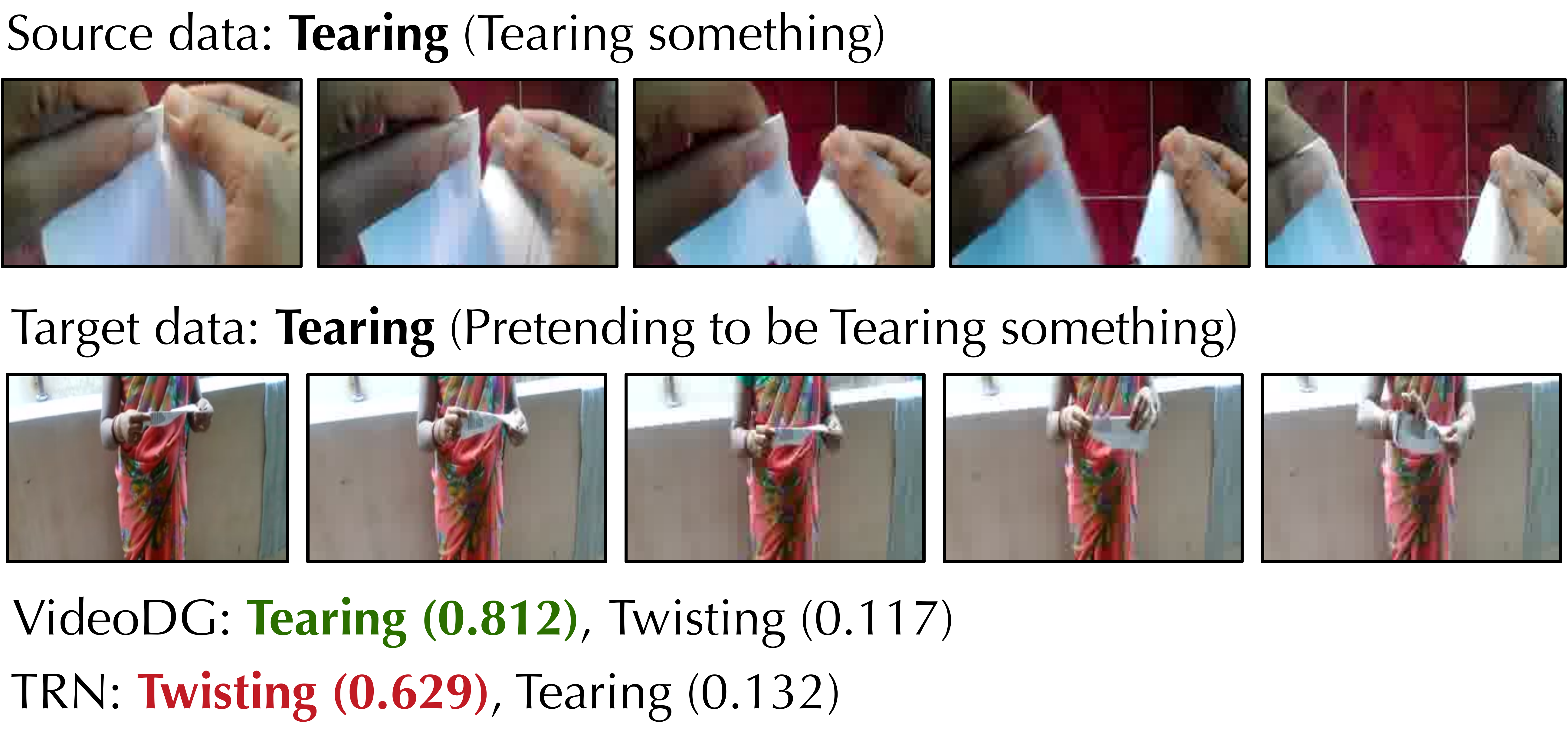}
    \vspace{-10pt}
    \caption{
    A showcase of the Something-Something benchmark. The first row shows training data from the source domain. The second row shows the test data from the target domains. 
    }
    \label{fig:sth_show}
\end{figure}

\begin{figure}[t]
\vspace{5pt}
    \centering
    \includegraphics[width=0.95\columnwidth]{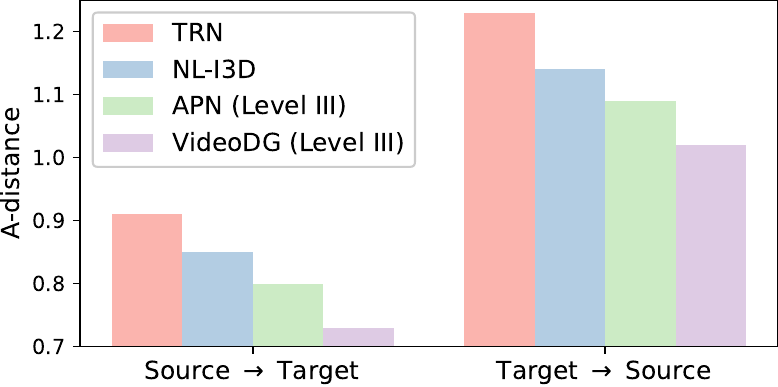}
    \vspace{-5pt}
    \caption{
    A-distances on the Something-Something benchmark that are obtained from features of TRN, NL-I3D, and APN (w/o or w/ RADA, as shown by the green and purple bars, respectively).
    }
    \label{fig:sth_a_distance}
\end{figure}

\begin{table*}[t]
\centering
\small
\caption{Selected categories from the Something-Something dataset to construct the cross-relation video domain generalization benchmark.}
\vspace{-5pt}
    \begin{tabular}{cll}
    \toprule
     Label & Source domain & Target domain \\
    \midrule
    0 & Attaching [something] to [something] & Trying but failing to attach [something] to [something]  \\
   	1 & Bending [something] until it breaks & Trying to bend [something unbendable] so nothing happens \\
   	2 & Wiping [something] off of [something] & Pretending or failing to wipe [something] off of [something] \\
	3 & Twisting [something] & Pretending or trying and failing to twist [something] \\
 	4 & Turning [something] upside down & Pretending to turn [something] upside down \\
 	5 & Picking [something] up & Pretending to pick [something] up \\
  	6 & Scooping [something] up with [something] & Pretending to scoop [something] up with [something] \\
 	7 & Squeezing [something] & Pretending to squeeze [something] \\
 	8 & Spreading [something] onto [something] & Pretending to spread air onto [something] \\
 	9 & Sprinkling [something] onto [something] & Pretending to sprinkle air onto [something] \\
 	10 & Taking [something] from [somewhere] & Pretending to take [something] from [somewhere] \\
 	11 & Tearing [something]  & Pretending to be tearing [something that is not tearable] \\
   	12 & Throwing [something] & Pretending to throw [something] \\
   	13 & Putting [something] behind [something] & Pretending to put [something] behind [something] \\
   	14 & Putting [something] underneath [something] & Pretending to put [something] underneath [something] \\
   	15 & Putting [something] into [something] & Pretending to put [something] into [something] \\
   	16 & Putting [something] onto [something] & Pretending to put [something] onto [something] \\
  	17 & Opening [something] & Pretending to open [something] without actually opening it \\
   	18 & Putting [something] on a surface & Pretending to put [something] on a surface \\
   	19 & Pouring [something] into [something] & Pretending to pour [something] out of [something] \\
   \bottomrule
    \end{tabular}
\label{Sth-category}
\end{table*}

\subsubsection{Main Results}
As shown in Table \ref{tab:sth}, due to the remarkable temporal domain shift, the classification accuracy remains low, and none of the compared models shows much generalizability on this challenging benchmark.
But still, the proposed VideoDG approach (\textit{i.e.}, APN trained with RADA) outperforms other models significantly.
Fig.~\ref{fig:sth_show} gives a showcase on this benchmark, where VideoDG correctly recognizes the action of \textit{tearing} even though there is a notable temporal domain shift. By contrast, the compared model misclassifies the action as \textit{twisting}, which is incorrect but reasonable since the person in the test video has not yet torn the thing apart. It is a very challenging case, considering the ambiguity of motion in sub-categories of \textit{twisting something} and \textit{pretending to tear something that is not tearable}. Furthermore, as shown in Fig.~\ref{fig:sth_a_distance}, VideoDG reduces the domain shift more effectively than TRN on the Something-Something benchmark.

The full list of the categories/sub-categories of the Something-Something benchmark is included in Table \ref{Sth-category}.

\section{Conclusion}

This paper introduced a new problem of video domain generalization, where models are trained on one source domain and evaluated on different unseen domains. We found that most video classification networks, even enhanced with existing methods for learning generalizable visual features, underperform in such settings due to the co-occurrence of spatial and temporal domain shifts. We proposed a novel method named VideoDG with two technical contributions. The first one is the new Adversarial Pyramid Network, which progressively learns generalizable and discriminative video representations at different pyramid levels. We then used the feature pyramid to generate adversarial examples in space-time, and thus derived the Robust Adversarial Domain Augmentation algorithm, which is the second contribution of this work. We constructed three video benchmarks with different kinds of spatial and temporal domain shifts, and validated the effectiveness of VideoDG on all benchmarks.

\section*{Acknowledgments}
We thank Xingqiang Du at Tsinghua for discussions.
This work was supported by the National Key R\&D Program of China (2020AAA0109201), NSFC grants (62022050, 62021002, 61772299), Beijing Nova Program (Z201100006820041), MOE Innovation Plan, and BNRist Innovation Fund.
Yunbo Wang is supported in part by Shanghai Sailing Program and CAAI-Huawei MindSpore Open Fund. Zhiyu Yao and Yunbo Wang contributed equally to this work.

\bibliographystyle{IEEEtran}
\bibliography{IEEEabrv,egbib}

\vspace{-10pt}
\begin{IEEEbiography}
[{\includegraphics[width=1in,height=1.25in,clip,keepaspectratio]{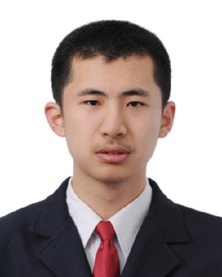}}]{Zhiyu Yao} received the BE degree in computer software from Tsinghua University, China, in 2019. He is working toward the PhD degree in computer software at Tsinghua University. His research interests include machine learning and computer vision.
\end{IEEEbiography}

\vspace{-10pt}
\begin{IEEEbiography}[{\includegraphics[width=1in,height=1.3in,clip,keepaspectratio]{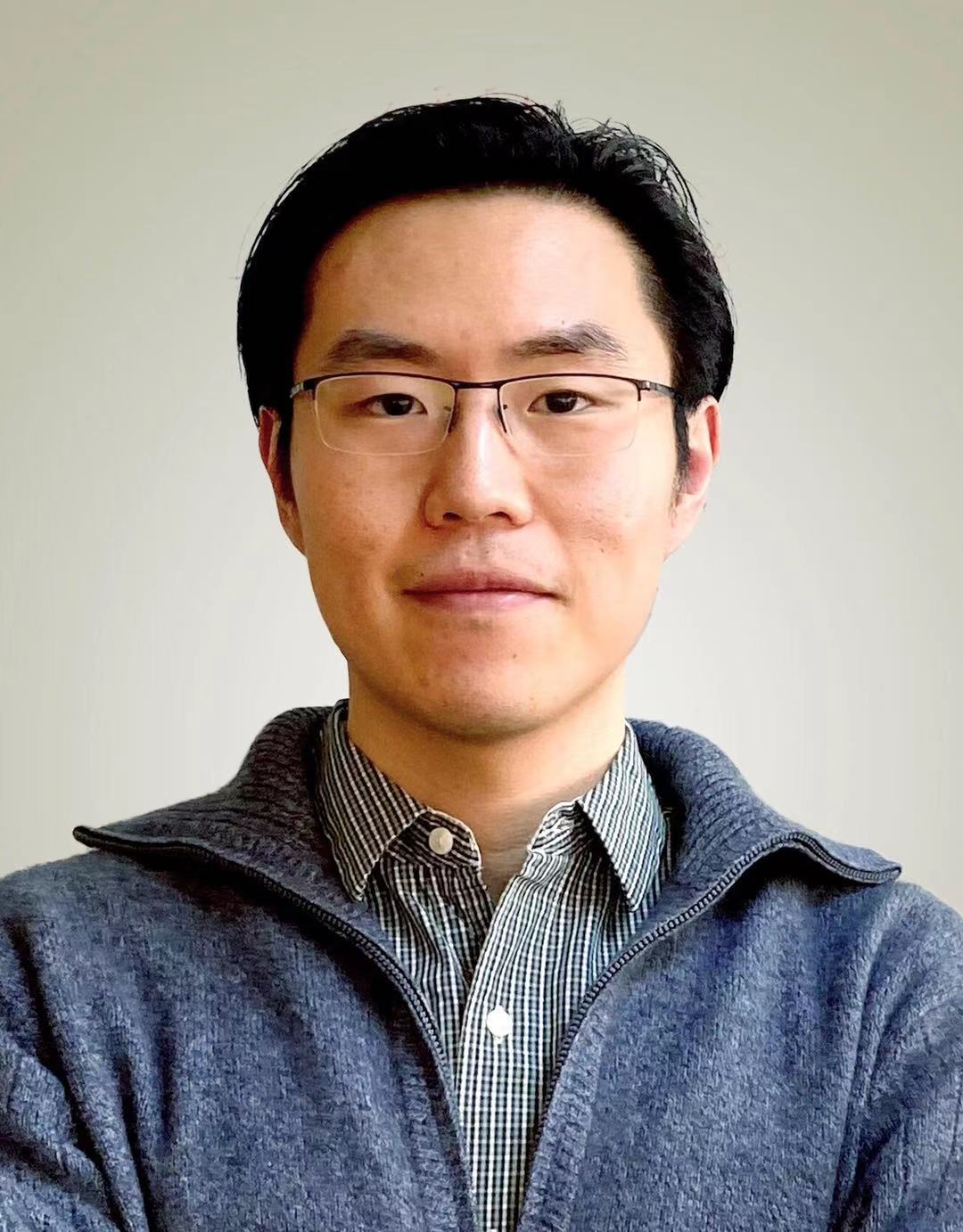}}]{Yunbo Wang} received the BE degree from Xi'an Jiaotong University in 2012, and the ME and PhD degrees from Tsinghua University in 2015 and 2020. He received the CCF Outstanding Doctoral Dissertation Award in 2020, advised by Philip S. Yu and Mingsheng Long. He is now an assistant professor at the AI Institute and the Department of Computer Science at Shanghai Jiao Tong University. He does research in deep learning, especially predictive learning, spatiotemporal modeling, and model-based decision making. 
\end{IEEEbiography}

\vspace{-10pt}
\begin{IEEEbiography}[{\includegraphics[width=1in,height=1.25in,clip,keepaspectratio]{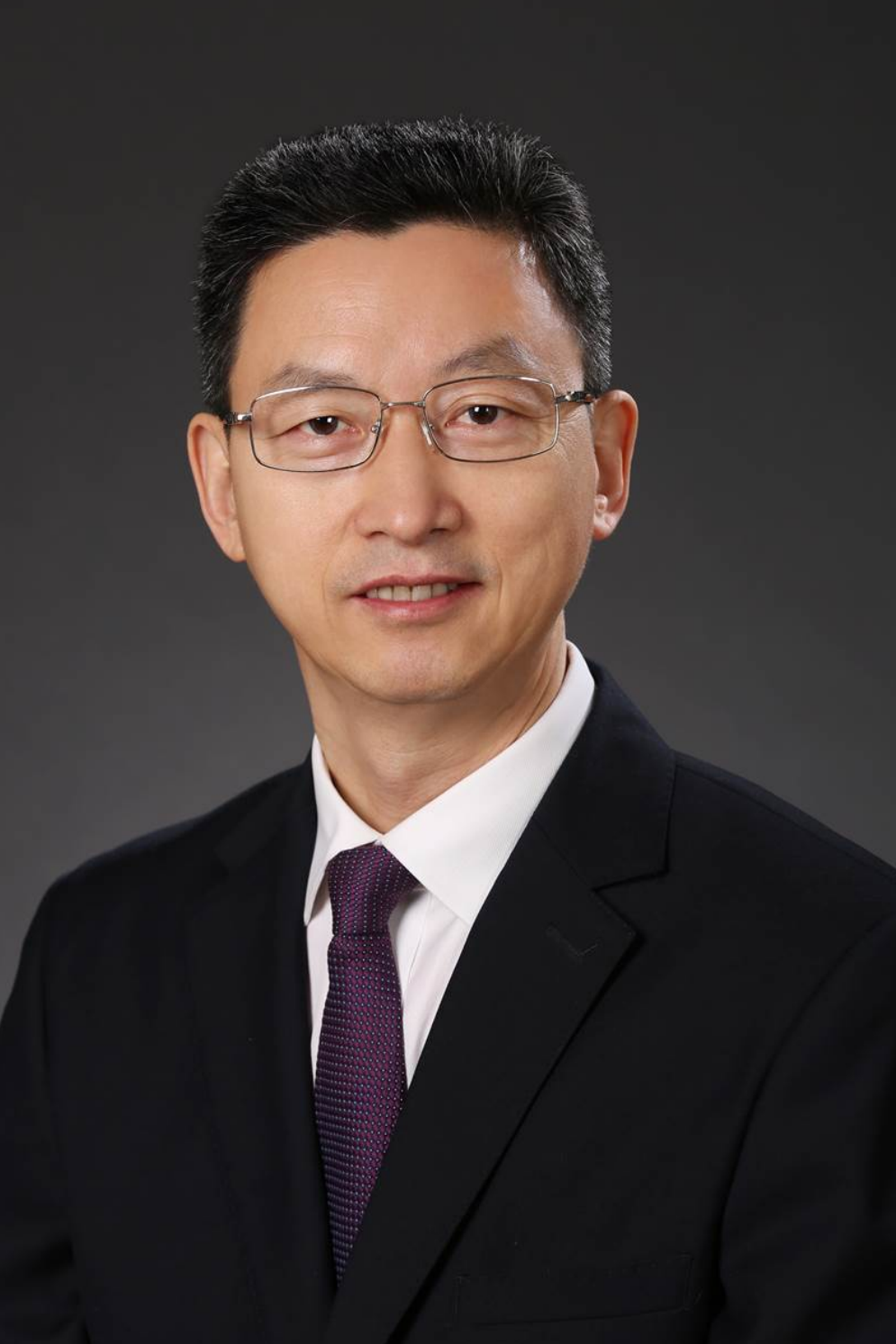}}]{Jianmin Wang} received the BE degree from Peking University, China, in 1990, and the ME and PhD degrees in computer software from Tsinghua University, China, in 1992 and 1995,
respectively. He is a full professor with the School
of Software, Tsinghua University. His research interests include big data management systems and large-scale data analytics. He led to develop a product data \& lifecycle management system, which has been deployed in hundreds of enterprises in China. He is leading to develop a big data platform in the National Engineering Lab for Big Data Software.
\end{IEEEbiography}

\vspace{-10pt}
\begin{IEEEbiography}[{\includegraphics[width=1in,height=1.25in,clip,keepaspectratio]{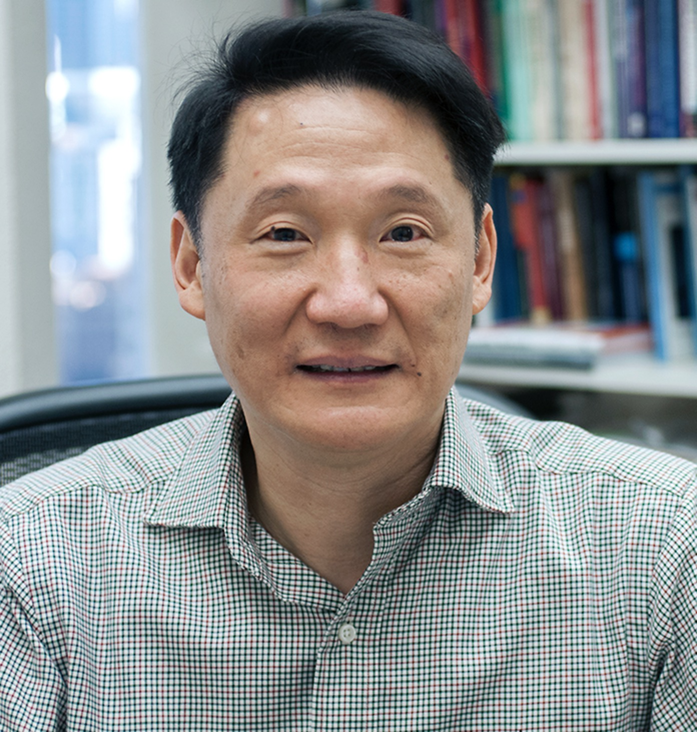}}]{Philip S. Yu} received his BS in electrical engineering from the National Taiwan University, and his MS and PhD also in electrical engineering from Stanford University in 1978. He is a Distinguished Professor at the University of Illinois at Chicago and Tsinghua University. Yu holds over 300 US patents, is ACM Fellow and IEEE Fellow, is Editor-in-Chief of ACM Transactions on Knowledge Discovery from Data, and has been awarded several awards by IBM and the IEEE. Yu's research interests are in the fields of data mining, social network, privacy preserving data publishing, data stream, database systems, and Internet applications and technologies. 
\end{IEEEbiography}

\vspace{-10pt}
\begin{IEEEbiography}[{\includegraphics[width=1in,height=1.25in,clip,keepaspectratio]{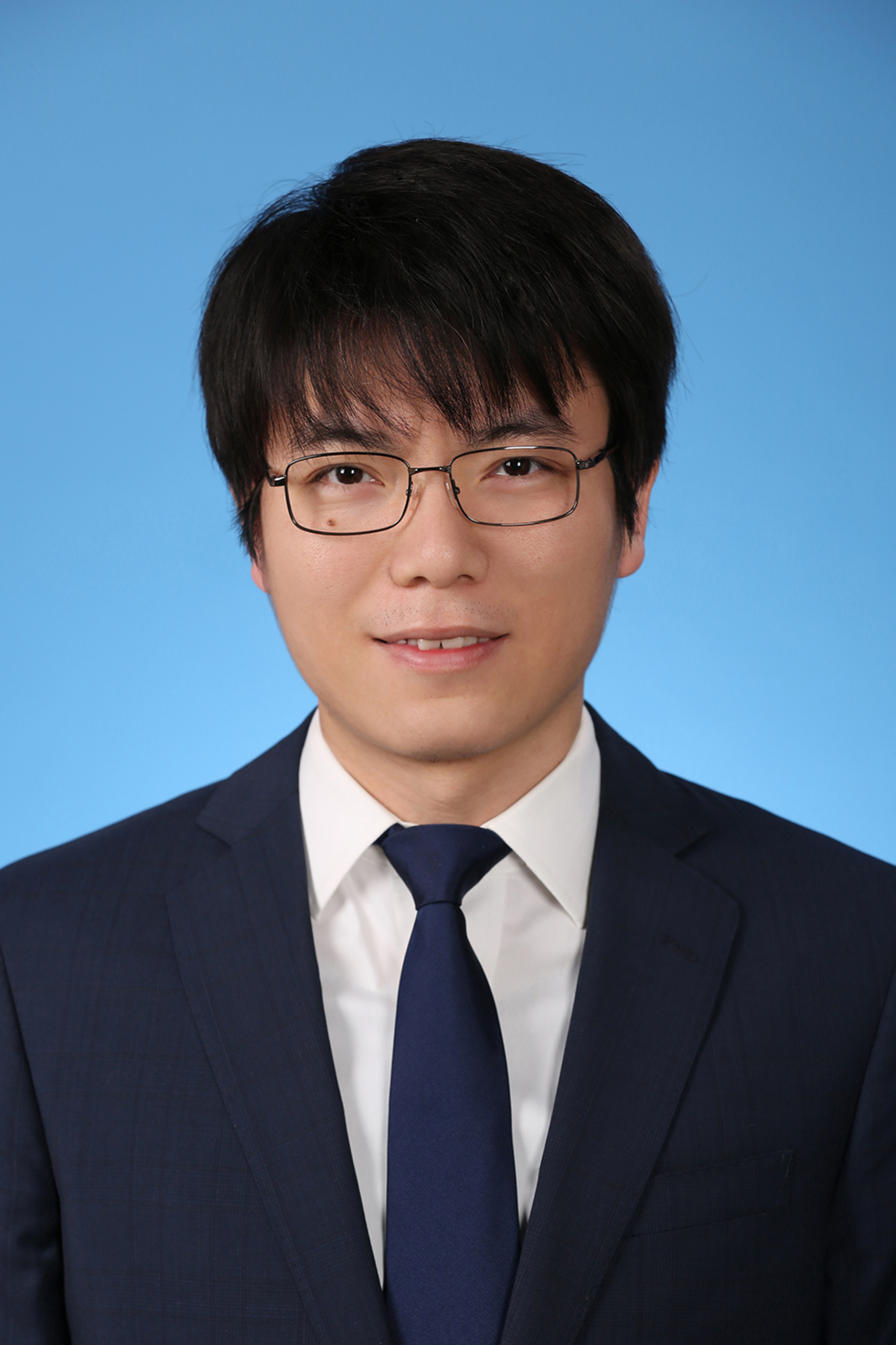}}]{Mingsheng Long} received the BE degree in electrical engineering and the PhD degree in computer science from Tsinghua University in 2008 and 2014 respectively. He is an associate professor with the School of Software, Tsinghua University. He was a visiting researcher in computer science, UC Berkeley from 2014 to 2015. He serves as Area Chairs of major machine learning conferences (ICML/NeurIPS/ICLR). His research is dedicated to theories and algorithms of machine learning, with special interests in transfer learning, deep learning, and learning with scientific knowledge.
\end{IEEEbiography}


\end{document}